\documentclass[10pt,journal,compsoc]{IEEEtran}
\ifCLASSOPTIONcompsoc
  \usepackage[nocompress]{cite}
\else
  \usepackage{cite}
\fi
\ifCLASSINFOpdf
\else
\fi

\usepackage{epsfig}
\usepackage{graphicx}
\usepackage{amsmath}
\usepackage{amssymb}
\usepackage{booktabs}
\usepackage{multirow}
\usepackage{adjustbox}
\usepackage[utf8x]{inputenc} 
\usepackage{csquotes}
\usepackage[table,xcdraw,dvipsnames]{xcolor}
\usepackage{nicefrac} 
\usepackage{pifont}
\usepackage{algorithmic}
\usepackage{geometry}
\usepackage[T1]{fontenc}
\usepackage[numbers]{natbib}
\usepackage[pagebackref=true,breaklinks=true,colorlinks,bookmarks=false,citecolor=blue,linkcolor=blue]{hyperref}

\definecolor{green}{rgb}{0.4, 0.69, 0.2}
\definecolor{purple}{rgb}{0.58, 0.44, 0.86}
\newcommand{\xmark}{\ding{55}}
\newcommand{\clswgan}{\texttt{f-CLSWGAN}}
\newcommand{\vaegan}{\texttt{f-VAEGAN}}

\newcommand{\highlight}[1]{\textcolor{blue}{#1}}
\def\redlozenge{\mathbin{\color{red}\blacklozenge}}
\def\bluelozenge{\mathbin{\color{blue}\blacklozenge}}
\def\yellowlozenge{\mathbin{\color{yellow}\blacklozenge}}
\def\greenlozenge{\mathbin{\color{green}\blacklozenge}}
\def\purplelozenge{\mathbin{\color{purple}\blacklozenge}}
\begin{document}

%%%%%%%%% TITLE
% \title{Latent Semantic Embedding and Feedback for Zero-Shot Recognition}
%FUSION-GAN: %Cross-Level Attribute
\title{Generative Multi-Label Zero-Shot Learning}
%\thispagestyle{empty}

% \author{Akshita Gupta*, Sanath Narayan*,Salman Khan,
% Fahad Shahbaz Khan, Ling Shao, Joost van de Weijer}%
% \IEEEcompsocitemizethanks{\IEEEcompsocthanksitem M. Shell was with the Department
% of Electrical and Computer Engineering, Georgia Institute of Technology, Atlanta,
% GA, 30332.\protect\\
% E-mail: see http://www.michaelshell.org/contact.html
% \IEEEcompsocthanksitem J. Doe and J. Doe are with Anonymous University.}% <-this % stops a space
% \thanks{Manuscript received April 19, 2005; revised August 26, 2015.}}

% $^1$Inception Institute of Artificial Intelligence, UAE \quad $^2$Mohamed Bin Zayed University of AI, UAE \\
% $^3$ Universitat Autònoma de Barcelona
% }
\author{Akshita Gupta*, Sanath Narayan*, Salman Khan, Fahad Shahbaz Khan, Ling Shao, Joost van de Weijer % <-this % stops a space
\IEEEcompsocitemizethanks{\IEEEcompsocthanksitem 
A. Gupta is with University of Guelph and Vector Institute. Correspondence to agupta22@uoguelph.ca \protect 
\IEEEcompsocthanksitem S. Narayan is with Technology Innovation Institute, UAE.
% S. Narayan are with the Inception Institute of Artificial Intelligence, UAE. 
\IEEEcompsocthanksitem S. Khan and F. S. Khan. are with Mohamed Bin Zayed University of AI, UAE. S. Khan is also with Australian National University, Australia. F. S. Khan is also with Link\"{o}ping University, Sweden.
\IEEEcompsocthanksitem L. Shao is with the UCAS-Terminus AI Lab, University of Chinese Academy of Sciences, Beijing, China.
 % L.Shao is with Terminus Group, China.
\IEEEcompsocthanksitem J. van de Weijer is with Computer Vision Center, Universitat Aut\`onoma de Barcelona, Spain. }}
% \IEEEcompsocthanksitem J. Doe and J. Doe are with Anonymous University.}% <-this % stops a space
% \thanks{Manuscript received April 19, 2005; revised August 26, 2015.}}

\markboth{IEEE Transactions on Pattern Analysis and Machine Intelligence}%
{Shell \MakeLowercase{\textit{et al.}}: Bare Demo of IEEEtran.cls for Computer Society Journals}

\IEEEtitleabstractindextext{%
%%%%%%%%% ABSTRACT
\begin{abstract}
 Multi-label zero-shot learning strives to classify images into multiple unseen categories for which no data is available during training. The test samples can additionally contain seen categories in the generalized variant. Existing approaches rely on learning either shared or label-specific attention from the seen classes. Nevertheless, computing reliable attention maps for unseen classes during inference in a multi-label setting is still a challenge. In contrast, state-of-the-art single-label generative adversarial network (GAN) based approaches learn to directly synthesize the class-specific visual features from the corresponding class attribute embeddings. However, synthesizing multi-label features from GANs is still unexplored in the context of zero-shot setting. 
%  In this work, we explore different strategies such as attribute fusion, feature fusion and cross-level attribute feature fusion for synthesizing multi-label visual features from their corresponding multi-label class embeddings.
When multiple objects occur jointly in a single image, a critical question is how to effectively fuse multi-class information.
In this work, we introduce different fusion approaches at the attribute-level, feature-level and cross-level (across attribute and feature-levels) for synthesizing multi-label features from their corresponding multi-label class embeddings.
To the best of our knowledge, our work is the first to tackle the problem of multi-label feature synthesis in the (generalized) zero-shot setting. Our cross-level fusion-based generative approach outperforms the state-of-the-art on three zero-shot benchmarks: NUS-WIDE, Open Images and MS COCO. Furthermore, we show the generalization capabilities of our fusion approach in the zero-shot detection task on MS COCO, achieving favorable performance against existing methods. Source code is available at \href{https://github.com/akshitac8/Generative_MLZSL}{https://github.com/akshitac8/Generative\_MLZSL}
%Comprehensive experiments are performed on three zero-shot classification benchmarks: NUS-WIDE, Open Images and MS COCO. Our cross-level fusion-based generative approach outperforms the state-of-the-art on all three datasets. Furthermore, we show the generalization capabilities of our fusion approach in the zero-shot detection task on MS COCO, achieving favorable performance against existing methods. 
%We also show the generalization capabilities of our approach in the zero-shot detection task on MS COCO. Our approach performs favorably in comparison to the state-of-the-art, achieving gains as high as 6.3\% mAP on NUS-WIDE.
\end{abstract}

}

\maketitle

\IEEEdisplaynontitleabstractindextext

% \fi
\IEEEpeerreviewmaketitle

\IEEEraisesectionheading{\section{Introduction}\label{sec:introduction}}

\IEEEPARstart{M}{ulti-label} classification is a challenging problem where the task is to recognize all labels in an image.
Typical examples of multi-label classification include, MS COCO~\cite{coco} and NUS-WIDE~\cite{nuswide} datasets, where an image may contain several different categories (labels). Most recent multi-label classification approaches address the problem by utilizing attention mechanisms~\cite{wang2017multi,yeattention,you2020cross}, recurrent neural networks~\cite{wang2016cnn,yazici2020orderless,nam2017maximizing}, graph CNNs~\cite{kipf2016semi,chen2019multi} and label correlations~\cite{weston2011wsabie,durand2019learning}. However, these approaches do not tackle the problem of multi-label zero-shot classification, where the task is to classify images into multiple new ``unseen'' categories at test time, without being given any corresponding visual example during the training. 
%instances for ``unseen'' classes are unavailable during training.
Different from zero-shot learning (ZSL), the test samples can belong to the seen or unseen classes in generalized zero-shot learning (GZSL). Here, we tackle the challenging problem of large-scale multi-label ZSL and GZSL.

Existing multi-label (G)ZSL approaches address the problem by utilizing global image representations~\cite{mensink2014costa,zhang2016fast}, structured knowledge graphs~\cite{lee2018multi} and attention-based mechanisms~\cite{huynh2020shared}. In contrast to the multi-label setting, single-label (generalized) zero-shot learning, where an image contains at most one category label, has received significant attention~\cite{jayaraman14nips,fu15pami,frome13nips,romera15icml,rohrbach13nips,Ye17cvpr,akata2015label,zsl-good-bad-ugly,xian2018feature,xian2019f}. State-of-the-art single-label (G)ZSL approaches~\cite{xian2018feature,Rafael18eccv,li19leveraging,huang19generative,Mandal19cvpr,xian2019f,narayan2020latent} are generative in nature. These approaches exploit the power of generative models, such as generative adversarial networks (GANs)~\cite{gan} and variational autoencoder (VAE)~\cite{kingma13iclr} to synthesize unseen class features. Typically, a feature synthesizing generator is utilized to construct single-label features. The generative approaches currently dominate single-label ZSL due to their ability to synthesize unseen class (fake) features by learning the underlying data distribution of seen classes (real) features. Nevertheless, the generator only synthesizes single-label features in the existing ZSL frameworks. To the best of our knowledge, the problem of designing a feature synthesizing generator for \textit{multi-label} ZSL paradigm is yet to be explored.

\begin{figure}[t]
\centering
\includegraphics[width=\columnwidth]{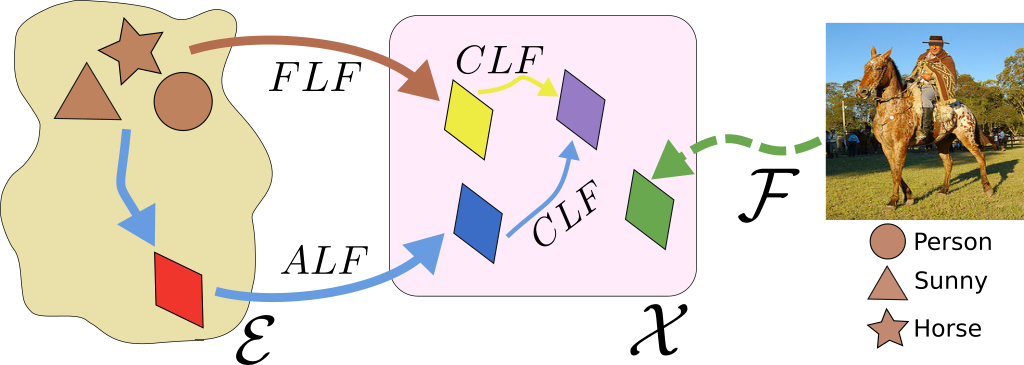}
\caption{A conceptual illustration of our three fusion approaches: attribute-level (ALF), feature-level (FLF) and cross-level feature (CLF), on an example image with three classes. Given an image, a feature extractor  $\mathcal{F}$ 
extracts real visual features ($\greenlozenge$) in $\mathcal{X}$.
ALF integrates the individual class-specific embeddings to generate a global image-level embedding ($\redlozenge$) in  $\mathcal{E}$. The image-level embedding is then used to synthesize the multi-label features ($\bluelozenge$). Different from ALF, the FLF synthesizes class-specific features and integrates them in the feature space $\mathcal{X}$. These integrated features are shown as $\yellowlozenge$. Our CLF takes as input synthesized features from ALF and FLF and enriches each respective feature by taking guidance from the other. The resulting enriched features are integrated to obtain the final image-level feature representation ($\purplelozenge$). The final image representation combines the advantage of label dependency of ALF and class-specific discriminability of FLF. The figure has been modified to include the color change suggested by the reviewers. 
\vspace{-0.1cm}
}
\label{fig:intro}
\end{figure}

%We introduce a generative approach for multi-label (generalized) zero-shot learning. 
In this work, we address the problem of multi-label (generalized) zero-shot learning by introducing an approach based on the generative paradigm. 
% 
%\highlight{When designing a generative multi-label zero-shot approach, an important issue to tackle is to effectively represent information from different multi-class objects occurring in an image. This is harder in comparison to single-label feature generation, where the discriminative information in an image is from a single class. Since \textit{multi-label} visual features representing all the different objects in an image need to be generated from a set of corresponding class attributes (embeddings), accurately fusing the information w.r.t each class embedding becomes critical.}
%Different from single-label feature generation, multi-label feature synthesis is desired to accurately integrate information regarding multiple objects jointly occurring in an image. 
% 
When designing a generative multi-label zero-shot approach, the main objective is to synthesize semantically consistent \textit{multi-label} visual features from their corresponding class attributes (embeddings). 
Multi-label visual features can be synthesized in two ways. (i) One approach is to integrate class-specific attribute embeddings at the input of the generator to produce a global image-level embedding vector. We call this approach attribute-level fusion (ALF). Here, the image-level embedding represents the holistic distribution of the positive labels in the image (see Fig.~\ref{fig:intro}). Since the generator in ALF performs global image-level feature generation, it is able to better capture label dependencies (correlations among the labels) in an image. However, such a feature generation has lower class-specific discriminability since the discriminative information with respect to the individual classes is not explicitly encoded.  (ii) A second approach is to synthesize the features from the class-specific embeddings individually and then integrate them in the visual feature space. We call this approach feature-level fusion (FLF), as shown in Fig.~\ref{fig:intro}. Although FLF better preserves the class-specific discriminative information in the synthesized features, it does not explicitly encode the label dependencies in an image due to synthesizing features independently of each other. This motivates us to investigate an alternative fusion approach for
multi-label feature synthesis.

In this work, we introduce an alternative fusion approach that combines the advantage of label dependency of ALF and the class-specific discriminability of FLF, during multi-label feature synthesis. We call this approach cross-level feature fusion (CLF), as in Fig.~\ref{fig:intro}. The CLF approach utilizes each individual-level feature and attends to the bi-level context (from ALF and FLF). As a result, individual-level features adapt themselves to produce enriched synthesized features, which are then pooled to obtain the CLF output. In  addition  to  multi-label  zero-shot  classification,  we investigate the proposed multi-label feature generation CLF approach for (generalized) zero-shot object detection.

% \subsection{Contributions}
\noindent \textbf{Contributions:} 
We propose a generative approach for multi-label (generalized) zero-shot learning. To the best of our knowledge, we are the first to explore the problem of multi-label feature synthesis in the zero-shot setting. 
We investigate three different fusion approaches (ALF, FLF and CLF) to synthesize multi-label features. Our CLF approach combines the  advantage  of  label  dependency of  ALF  and  the  class-specific discriminability of FLF. Further, we integrate our fusion approaches into two representative generative architectures: \clswgan{}~\cite{xian2018feature} and \vaegan{}~\cite{xian2019f}. We hope our simple and effective method serves as a solid baseline and aids ease future research in generative \textit{multi-label} zero-shot learning.

We evaluate our (generalized) zero-shot classification approach on  NUS-WIDE~\cite{nuswide}, Open Images~\cite{openimages} and MS COCO~\cite{coco}. Our CLF approach achieves consistent improvement in performance over both ALF and FLF. Furthermore, CLF outperforms state-of-the-art methods on \textit{all} datasets. On the large-scale Open Images dataset, our CLF achieves absolute gains of $18.7\%$ and $17.6\%$ over the state-of-the-art in terms of GZSL F1 score at \textit{top}-$K$ predictions, $K \in \{10,20\} $, respectively.
We additionally evaluate CLF for (generalized) zero-shot object detection, achieving favorable results against existing methods on MS COCO.

% In addition to classification, we evaluate CLF for (generalized) zero-shot object detection, achieving favorable results against existing methods on MS COCO.

\section{\hspace{-0.3em}Generative Single-Label Zero-Shot Learning\label{sec:slzsl}}
As discussed earlier, state-of-the-art single-label zero-shot approaches~\cite{xian2018feature,Rafael18eccv,li19leveraging,huang19generative,Mandal19cvpr,xian2019f} are generative in nature, utilizing the power of generative models (\textit{e.g.}, GAN \cite{gan}, VAE~\cite{kingma13iclr}) to synthesize unseen class features. 
Here, each image is assumed to have a \textit{single} object category label (\textit{e.g.}, CUB~\cite{cub}, FLO~\cite{flo} and AWA~\cite{zsl-good-bad-ugly}).
%generative approaches~\cite{xian2018feature,xian2019f} have been investigated for multi-class zero-shot learning (ZSL), where each image is assumed to have a \textit{single} object category label (\textit{e.g.}, CUB~\cite{cub}, FLO~\cite{flo} and AWA~\cite{awa}).
%In the single-label multi-class ZSL, the goal is to classify images into new unseen classes, which are unknown during the training stage. 
%Here, an image contains only \textit{one} of the seen (unseen) classes during training (test). 
% 
% \highlight{In the ZSL setting, the goal is to classify images into new unseen classes, which are unknown during the training stage.} % Repetition or required for comparison with multi-label?
% Furthermore, the test samples can belong to seen or unseen classes in the generalized variant (GZSL), thereby making it a harder problem due to the classifier bias towards the seen classes. % Both ZSL and GZSL settings involve training and testing with images containing \textit{single} object categories (\textit{e.g.}, CUB~\cite{cub}, FLO~\cite{flo} and AWA~\cite{awa}).  

% 
\noindent\textbf{Problem Formulation:}
Let $x \in \mathcal{X}$ denote the encoded feature instances of images and $y \in \mathcal{Y}^s$ the corresponding class labels from the set of $S$ seen class labels $\mathcal{Y}^s = \{y_1,\ldots,y_S\}$. Let $\mathcal{Y}^u = \{y_{S+1},\ldots,y_{C}\}$ denote the set of $U$ unseen classes, which is disjoint from the seen classes $\mathcal{Y}^s$. Here, the total number of seen and unseen classes is denoted by $C{=}S{+}U$. The relationships among all the seen and unseen classes are described by the category-specific semantic embeddings $e(k) \in \mathcal{E}$, $\forall k \in \mathcal{Y}^s \cup \mathcal{Y}^u$. 
To learn the ZSL and GZSL classifiers, existing single-label GAN-based approaches~\cite{xian2018feature,li19leveraging,huang19generative,xian2019f} first learn a generator using the seen class features $x_s$ and corresponding class embeddings $e(y_s)$. Then, the learned generator and the unseen class embeddings $e(y_u)$ are used to synthesize the unseen class features $\tilde{x}_u$. 
%to synthesize the features using the seen class features $x_s$ and corresponding embeddings $e(y_s)$. Then, the unseen class features $\tilde{x}_u$ are synthesized using the learned model and the unseen class embeddings $e(y_u)$. 
The resulting synthesized features $\tilde{x}_u$, along with the real seen class features $x_s$, are further deployed to train the final classifiers $f_{z}\!:\! \mathcal{X} {\rightarrow} \mathcal{Y}^U$ and $f_{gz}\!:\! \mathcal{X} {\rightarrow} \mathcal{Y}^C$. 
%Next, we briefly describe the feature synthesis stage employed in existing single-label GAN-based zero-shot frameworks.
% , to generate unseen class features. 

Typically, GAN-based single-label zero-shot frameworks utilize a feature synthesizing generator $G\!:\!\mathcal{Z}{\times}\mathcal{E} {\rightarrow} \mathcal{X}$ and a discriminator $D$. Both $G$ and $D$ compete against each other in a two player minimax game. While $D$ attempts to accurately distinguish real image features $x$ from generated features $\tilde{x}$, $G$ attempts to fool $D$ by generating features that are semantically close to real features. Since class-specific features are to be synthesized, a conditional Wasserstein GAN~\cite{wgan} is employed due to its more stable training, by conditioning both $G$ and $D$ on the embeddings $e(y)$. Here, $G$ learns to synthesize class-specific features $\tilde{x}$ from the corresponding single-label embeddings $e(y)$, given by $\tilde{x}\!=\!G(z, e(y))$. Nevertheless, the generator only synthesizes single-label features in existing zero-shot learning frameworks. To the best of our knowledge, the problem of designing a feature synthesizing generator for the multi-label zero-shot learning paradigm is yet to be investigated.

\section{Generative Multi-Label Zero-Shot Learning\label{sec:method}}
% 1. Definition of multi-label zero-shot task, with symbols 
% 2. Preliminaries: f-CLSWGAN for single label. Conclude with the intuitive points when designing ML ZSL arch. Small conceptual fig
% 3. Motivation for EF, LF, Hybrid - Complementary characteristics/properties.
% 4. The actual EF, LF and hybrid.
% 5. f-vaegan?
% 6. Extension to ml zs detection

% We present an approach, \proposed, for multi-label zero-shot learning (ML-ZSL). Unlike the single-label ZSL, images at test time contain multiple unseen classes in ML-ZSL. (\textit{e.g.}, MS COCO~\cite{coco}, NUS-WIDE~\cite{nuswide} and Open Images~\cite{openimages}).

As discussed earlier, most real-world tasks involve multi-label recognition, where an image can contain multiple and wide range of category labels (\textit{e.g.}, Open Images~\cite{openimages}). Multi-label classification becomes more challenging in the zero-shot learning setting, where the test set either contains only unseen classes (ZSL) or both seen and unseen classes (GZSL). 
% As discussed earlier, most real-world tasks involve multi-label recognition, where an image can contain multiple and wide range of category labels (\textit{e.g.}, MS COCO~\cite{coco}, NUS-WIDE~\cite{nuswide} and Open Images~\cite{openimages}). The multi-label classification problem becomes more challenging in the zero-shot learning setting, where the test set either contains only unseen classes (ZSL) or both seen and unseen classes (GZSL). 
% Existing approaches address the multi-label zero-shot learning problem by utilizing attention-based mechanisms~\cite{huynh2020shared}, structured knowledge graphs~\cite{lee2018multi} and global image representations~\cite{zhang2016fast,mensink2014costa}. 
In this work, we propose a generative multi-label zero-shot learning approach that exploits the capabilities of generative models to learn the underlying data distribution of seen classes. This helps to mimic the fully-supervised setting by synthesizing (fake) features for unseen classes. 
%While generative approaches have been extensively studied for single-label zero-shot learning, we are the first to address the problem of multi-label feature synthesis in the multi-label (generalized) zero-shot setting.
%\highlight{When designing a generative multi-label zero-shot approach, an important issue to tackle is to effectively represent information from different multi-class objects occurring in an image. This is harder in comparison to single-label feature generation, where the discriminative information in an image is from a single class. Since \textit{multi-label} visual features representing all the different objects in an image need to be generated from a set of corresponding class attributes (embeddings), accurately fusing the information w.r.t each class embedding becomes critical.}
Different from the single-label setting, multi-label feature synthesis requires accurate integration of multi-class information regarding different objects jointly occurring in an image. Here, \textit{multi-label} visual features representing different objects in an image need to be synthesized from a set of corresponding class embeddings. Thus, accurately fusing class information w.r.t each label embedding becomes critical in multi-label feature synthesis.
% Next, we describe the problem formulation of generative multi-label (generalized) zero-shot learning. 

%While generative approaches currently constitute the state-of-the-art for single-label zero-shot learning, we are the first to propose a generative approach for multi-label zero-shot learning. 
%since generative models possess the capability to learn the underlying the data distribution of seen classes. This helps to mimic the fully supervised settings by synthesizing (fake) features for unseen classes. 
%While generative approaches currently constitute the state-of-the-art for single-label zero-shot learning, they are yet to be investigated for multi-label zero-shot learning. 

%Despite their recent success on single-label zero-shot learning, generative approaches are yet 

%Unlike the single-label paradigm, where a single image contains a single object, images in multi-label setting contain multiple objects. While test samples only contain multiple unseen classes in multi-label ZSL, test samples can contain multiple seen and unseen classes in the ML-GZSL setting.

% 
\noindent\textbf{Problem Formulation:} 
In contrast to single-label zero-shot learning, here, $x \in \mathcal{X}$ denotes the encoded feature instances of multi-label images and $y \in \{0,1\}^S$ the corresponding multi-hot labels from the set of $S$ seen class labels $\mathcal{Y}^s$. Let $x$ denote the multi-label feature instance of an image and $y$, its multi-hot label with $n$ positive classes in the image. Then, the set of attributes for the image can be denoted as $e(y) {=} \{e(y_{j}), \forall j\!:\!y[j]{=}1 \}$, where $|e(y)|{=}n$. Here, we use GloVe~\cite{pennington2014glove} vectors of the class names as the attributes $e(y_{j})$, as in~\cite{huynh2020shared}.
Now, the generator's task is to learn to synthesize the multi-label features $x$ from the associated embedding vectors $e(y)$.
Post-training of $G$, multi-label features corresponding to the unseen classes are synthesized. The resulting synthesized features along with the real seen class features are deployed to train the final classifiers $f_{z}\!:\!\mathcal{X} {\rightarrow} \{0,1\}^U$ and $f_{gz}\!:\! \mathcal{X} {\rightarrow} \{0,1\}^C$. %Next, we investigate three different approaches to synthesize unseen multi-label features. 

\begin{figure*}[t]
    \centering
    \includegraphics[width=0.87\textwidth]{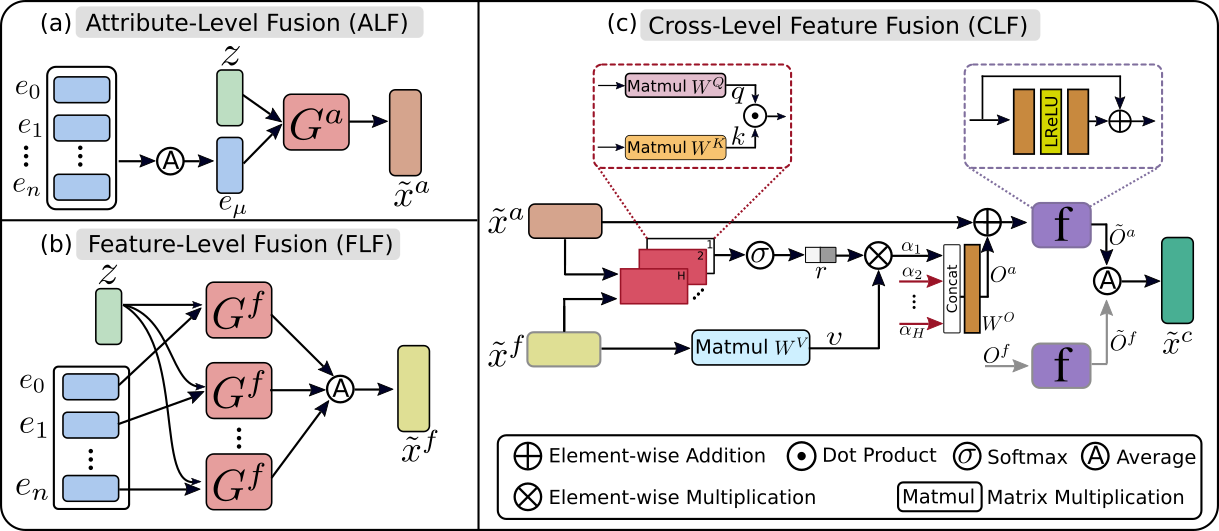}
    \caption{Overview of three different approaches to synthesize unseen multi-label features. The attribute-level fusion (ALF) generates a global image-level embedding vector from a set of class-specific embedding vectors corresponding to multiple labels in an image (Sec.~\ref{sec:att_fuse}). In ALF, the generator synthesizes global features $\tilde{x}^a$ that capture the correlations among the labels in the image. On the other hand, the feature-level fusion (FLF) synthesizes features from individual class-specific embeddings (Sec.~\ref{sec:feat_fuse}). As a result, the generator produces class-specific latent features which are then integrated to obtain synthesized features $\tilde{x}^f$. The cross-level fusion (CLF) combines the advantages of ALF and FLF, during feature generation (Sec.~\ref{sec:cross_fuse}). Specifically, it uses each individual-level feature ($\tilde{x}^a, \tilde{x}^f$) to attend to the bi-level context and adapt itself to generate $\tilde{o}^a, \tilde{o}^f$. These enriched features are then pooled to obtain the CLF output, which represents our final synthesized feature $\tilde{x}^c$.\vspace{-0.2cm} % $\tilde{x}^a$ and $\tilde{x}^f$ are projected onto a low-dimensional space to construct query-key-value triplets ($q$, $k$, $v$) using a total of $H$ projection heads. 
    }
    \label{fig:caf_fuse}
\end{figure*}

\subsection{Generative Multi-Label Feature Synthesis}
%As discussed earlier, existing GAN-based ZSL approaches (cite), including \clswgan{}~\cite{xian2018feature}, tackle the problem in single-label settings, where a label $y_i$ of a feature instance $\tilde{x}_i$ can belong to one of the classes in $\mathcal{Y}^s \cup \mathcal{Y}^u $. 
% 
%The multi-label features can be synthesized  

To synthesize multi-label features, we introduce three fusion approaches: attribute-level fusion (ALF), feature-level fusion (FLF) and cross-level feature fusion (CLF).

\subsubsection{Attribute-Level Fusion\label{sec:att_fuse}}

% In the case of multi-label zero-shot setting, an image comprises multiple object classes. Since each class is described by an attribute embedding vector, an image with multiple labels, in turn, can be described by the corresponding embedding vectors. Let $x_i$ denote the feature instance of the $i^{th}$ image and $y_i$, its multi-hot label with $n_i$ positive classes in the image. With a slight abuse of notation, the set of attribute embeddings for the $i^{th}$ image can be denoted as $e(y_i) = \{e(y_{ij}), \forall j : y_i[j]=1 \}$, where $|e(y_i)|=n_i$. Now, the generator's task is to learn to synthesize the multi-label features $x_i$ from the associated embedding vectors $e(y_i)$. 

% \noindent\textbf{Attribute-Level Fusion:} 
In attribute-level fusion (ALF) approach, a global image-level embedding vector is obtained from a set of class-specific embedding vectors that correspond to multiple labels in the image. The image-level embedding represents the global description of the positive labels in an image.  The global embedding 
$e_\mu$ is obtained by averaging the individual class embeddings $e(y)$. This embedding $e_\mu$ is then input to the generator $G^a$ along with the noise $z$ for synthesizing the feature $\tilde{x}^a$. The attribute-level fusion is then given by
%For simplicity, we choose the average operation for obtaining the global embedding $e_i$ from the set of individual embeddings $e(y_i)$.
%Given the set of embedding vectors $e(y_i)$, a global embedding vector $e_i$ is obtained. The resulting embedding $e_i$ represents the  description of the positive classes in image $i$, at a global-level. For simplicity, we choose the average operation for obtaining the global embedding $e_i$ from the set of individual embeddings $e(y_i)$. Next, the resulting global embedding $e_i$ is input to the generator $G^a$ along with the noise $z$ for synthesizing the feature $\tilde{x}_i^a$. The attribute-level fusion is then given by,
% 
% 
\begin{equation}
  \label{eq:att_fuse}
  e_\mu =  \frac{1}{n}\sum_{j:y[j]=1} e(y_{j}) \quad \text{and}  \quad 
  \tilde{x}^a = G^a(z,e_\mu).
\end{equation}
Fig.~\ref{fig:caf_fuse}\highlight{(a)} shows the feature synthesis of $\tilde{x}^a$ from the embedding $e_\mu$.~The generator $G^a$ in ALF performs global image-level feature generation, thereby capturing label dependencies (correlations among labels) in an image. However, such a generation from the global embedding has lower class-specific discriminability since it does not explicitly encode discriminative information w.r.t individual classes. 
% \highlight{Further, it is restrictive when dealing with a large number of labels.}

%is likely to miss discriminative information with respect to each class, thereby lacking class-specific discriminability. 

%restrictive when dealing with a large number of labels (see Fig toy). 

%the global contextual information with respect to the labels in the image. For instance, an image comprising person and vehicle categories is likely to contain a road. In this case, the global feature generation is able to exploit the additional information (\textit{e.g.}, road) at the image-level, despite such information not being explicitly designated as an object class. 

%label correlation in an image. 

%which accepts a semantic map containing several
%object classes and aims to generate the appearance of all the
%different classes

% However, since the feature is generated from a global embedding, $\tilde{x}_i^a$ is likely to miss class-specific discriminative information at a local level. \highlight{Is it ending negatively?}

\subsubsection{Feature-Level Fusion}\label{sec:feat_fuse}
Here, we introduce a feature-level fusion (FLF) approach,
% To address the problems induced by the ALF approach, 
which synthesizes the features from the class-specific embeddings individually. This allows the FLF approach to better preserve the class-specific discriminability in the synthesized features.
%In this way, the class-specific discriminative information in the synthesized features is retained. 
Different from ALF that integrates the class embeddings, the FLF first inputs the $n$ class-specific embeddings $e(y_{j})$ to $G^f$ and generates $n$ class-specific latent features $\tilde{x}_{j}$. These features are then integrated through an average operation to obtain the final synthesized feature $\tilde{x}^f$. The feature-level fusion (FLF) is denoted by
\begin{equation}
  \label{eq:feat_fuse}
  \tilde{x}_{j} = G^f(z,e(y_j)) \quad \text{and} \quad \tilde{x}^f = \frac{1}{n}\sum_{j:y[j]=1} \tilde{x}_{j}.
\end{equation}
Fig.~\ref{fig:caf_fuse}\highlight{(b)} shows the feature generation of $\tilde{x}^f$ from the individual embeddings $e(y_{j})$. We observe from Eq.~\ref{eq:feat_fuse} that for a fixed noise $z$, the generator $G^f$ synthesizes a fixed latent feature $\tilde{x}_{j}$ for class $j$, regardless of the presence/absence of other classes in an image.
Thus, while the generated latent features $\tilde{x}_{j}$ better preserve class-specific discriminative information of the positive classes $j$ present in the image, $G^f$  synthesizes them independently of each other.
% While the generated latent features $\tilde{x}_{ij}$ have class-specific discriminative information of the positive classes $j$ present in the image, they are synthesized independent of each other. 
As a result, the synthesized feature $\tilde{x}^f$ does not explicitly encode the label dependencies in an image. 

%global discriminative information, which is crucial for enhanced feature synthesis. 

As discussed above, both the aforementioned fusion approaches (ALF and FLF) have shortcomings when synthesizing multi-label features. Next, we introduce a fusion approach that combines the advantage of the label dependency of ALF and class-specific discriminability of FLF.

\subsubsection{Cross-Level Feature Fusion}
\label{sec:cross_fuse}
The proposed cross-level feature fusion (CLF) aims to combine the advantages of both ALF and FLF. The CLF approach (see Fig.~\ref{fig:caf_fuse}\highlight{(c)}) incorporates label dependency and class-specific discriminability in the feature generation stage as in the ALF and FLF, respectively. 
%Given the complimentary features obtained from attribute-level and feature-level fusion blocks, we aim to develop an information consolidation pipeline to enhance the learned representations. 
To this end, $\tilde{x}^f$ and $\tilde{x}^a$ are forwarded to a feature fusion block. Inspired by the multi-headed self-attention \cite{vaswani2017attention}, the feature fusion block enriches each respective feature by taking guidance from the other branch. Specifically, we create a matrix $\tilde{x} \in \mathbb{R}^{2 \times d}$ by stacking the individual features $\tilde{x}^f$ and $\tilde{x}^a$. Then,  these features are linearly projected to a low-dimensional space ($d' = \nicefrac{d}{H}$) to create query-key-value triplets using a total of $H$ projection heads,
$$ {q}_h = \tilde{x}  {W}_h^Q, k_h =  \tilde{x}  {W}_h^K, v_h =  \tilde{x}  {W}_h^V, $$
where $ {W}_h^Q , {W}_h^K , {W}_h^V \in \mathbb{R}^{d \times d'}$ and  $h \in \{1,2, .. , H\}$. For each feature, a status of its current form is kept in the `\emph{value}' embedding, while the \emph{query} vector derived from each input feature is used to find its correlation with the \emph{keys} obtained from both the features, as we elaborate below.

\begin{figure}
  \centering
  \begin{minipage}[c]{0.4\columnwidth}
    \includegraphics{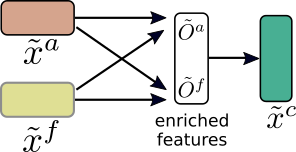}
  \end{minipage}\quad
  \begin{minipage}[c]{0.5\columnwidth}
    \caption{Conceptual illustration showing the cross-level fusion mechanism. The multi-label feature $\tilde{x}^a$ that is generated by ALF gets enriched by employing attention on itself and $\tilde{x}^f$ from FLF. Similarly, $\tilde{x}^f$ also gets enriched with the aid of $\tilde{x}^a$. The resulting enriched features $\tilde{o}^a$ and $\tilde{o}^f$ are then combined to obtain the final CLF feature $\tilde{x}^c$.}
    \label{fig:clff_conceptual}
  \end{minipage}
\end{figure}

Given these triplets from each head, the features undergo two levels of processing (i) intra-head processing on the triplet and (ii) cross-head processing. For the first case, the following equation is used to relate each query vector with `\emph{keys}' derived from both the features. The resulting normalized relation scores ($r_h \in \mathbb{R}^{2 \times 2}$) from the softmax function ($\sigma$) are used to reweight the corresponding value vectors, thereby obtaining the attended features $\alpha_h  \in \mathbb{R}^{2 \times d'}$, 
$$ \alpha_h = r_h v_h , \quad \text{ where }\; r_h = \sigma ( \frac{q_h k^{\top}_h }{\sqrt{d'}} ).$$
To aggregate information across all heads, these attended low-dimensional features from each head are concatenated and processed by an output layer to generate the original $d$-dimensional output vectors $o \in \mathbb{R}^{2 \times d}$, 
\begin{align}
    o = [\alpha_1 ; \alpha_2 ; \ldots \alpha_H] W^O, %\quad \text{ s.t., } W^O \in \mathbb{R}^{D \times D},
\end{align}
where $W^O \in \mathbb{R}^{d \times d}$ is a learnable weight matrix. 
After obtaining the self-attended features $o$, a residual branch is added from the input to the attended features  and further processed with a small residual sub-network $f(\cdot)$ to help the network first focus on the local neighbourhood and then progressively pay attention to the other-level features, 
\begin{equation}
\tilde{o} = f(\tilde{x} + o) + (\tilde{x} + o), \quad \text{s.t., } \tilde{o} \in \mathbb{R}^{2\times d}. 
\end{equation}
This encourages the network to selectively focus on adding complimentary information to the source vectors $\tilde{x}$. 
Finally, we mean-pool the matrix $\tilde{o} $ along the row dimension to obtain a single cross-level fused feature $\tilde{x}^{c} \in \mathbb{R}^{d}$, 
\begin{equation}
 \label{eq:clf_out}
  \tilde{x}^{c} = \frac{1}{2} \sum_{j \in \{1,2 \}} \tilde{o}^{j,k}.   
\end{equation}
The cross-level fused feature $\tilde{x}^{c}$ is obtained by effectively fusing the features generated from ALF and FLF (see Fig.~\ref{fig:clff_conceptual}). As a result, $\tilde{x}^{c}$ explicitly encodes the correlation among labels in the image, in addition to the class-specific discriminative information of the positive classes present. Next, we describe the integration of our CLF in representative generative architectures for multi-label zero-shot classification.

% The cross-level fused feature $\tilde{x}^{c}$ is obtained by effectively fusing the features generated from ALF and FLF. As a result, $\tilde{x}^{c}$ explicitly encodes the correlation among labels in the image, in addition to the class-specific discriminative information of the positive classes present in the image. Next, we describe the integration of our CLF approach in two representative generative architectures for multi-label (generalized) zero-shot classification. 

% \begin{figure}[t]
% \centering
% \includegraphics[width=0.8\columnwidth]{images/plot2.pdf}
% \caption{test seen-unseen }
% \end{figure}

\subsection{Multi-Label Zero-Shot Classification\label{sec:feat_syn}}
We integrate our fusion approaches in two representative generative architectures (\clswgan{}~\cite{xian2018feature} and \vaegan{}~\cite{xian2019f}) for  multi-label zero-shot classification. Both \clswgan{} and \vaegan{} have been shown to achieve promising performance for single-label zero-shot classification. Since CLF integrates both ALF and FLF, we only describe here the integration of CLF in the two classification frameworks. Next, we describe the integration of CLF in \clswgan{}, followed by \vaegan{}.

%The overall architecture of the proposed \proposed{}, based on a \clswgan{}, is shown in Fig.~\ref{fig:overall_arch}. 

% #######################
%\noindent\textbf{\texttt{f-CLSWGAN}:} 
%Here, we first describe the single-label zero-shot GAN-based approach. 
Briefly, \clswgan{} comprises a WGAN, conditioned on the embeddings $e(y)$, and a seen class classifier $f_{cls}$. We replace the standard generator of \clswgan{} with our multi-label feature generator (CLF) to synthesize multi-label features $\tilde{x}^{c}$. The resulting multi-label WGAN loss is, %given by}
% Briefly, \clswgan{} comprises a conditional WGAN (conditioned on the embeddings $e(y)$) and a seen class classifier $f_{cls}$. Here, we replace the standard generator of the \clswgan{} with our multi-label feature generation (CLF) to synthesize multi-label features $\tilde{x}^{c}$. The resulting multi-label WGAN loss is given by
% 
% 
\begin{align}
\label{eq:wgan}
\mathcal{L}_{W} = &~\mathbb{E}[D(x_s,e(y_s))] - \mathbb{E}[D(\tilde{x}_s^{c},e(y_s))] -  \\ 
                  & \lambda \mathbb{E}[\left(||\nabla_{\hat{x}} D(\hat{x},e(y_s))||_2 - 1\right)^2], \nonumber
\end{align}
where $\tilde{x}_s^{c}$ is synthesized using Eq.~\ref{eq:clf_out} for the seen classes,
% $\tilde{x}_s^{c}\!=\!\text{CLF}(G^a(z,e(y_s),G^f(z,e(y_s))$ is a generated feature from seen classes,
$\lambda$ is the penalty coefficient and $\hat{x}$ is a convex combination of $x_s$ and $\tilde{x}_s^{c}$. Furthermore, a classifier $f_{cls}$, trained on the seen classes, is employed to encourage the generator to synthesize features that are well suited for final ZSL/GZSL classification. The final objective for training our CLF-based \clswgan{} in a multi-label setting is given by,
\begin{equation}
    \label{eq:clswgan}
    \min_G \max_D \mathcal{L}_W + \alpha \mathbb{E}[\text{BCE}(f_{cls}(\tilde{x}^c_s, y_s))],
\end{equation}
where $\text{BCE}(f_{cls}(\tilde{x}^c_s), y_s)$ denotes the standard binary cross entropy loss between the predicted multi-label $f_{cls}(\tilde{x}_s)$ and the ground-truth multi-label $y_s$ of feature $x_s$.

% ##########################

\begin{figure}
    \centering
    \includegraphics[width=\columnwidth]{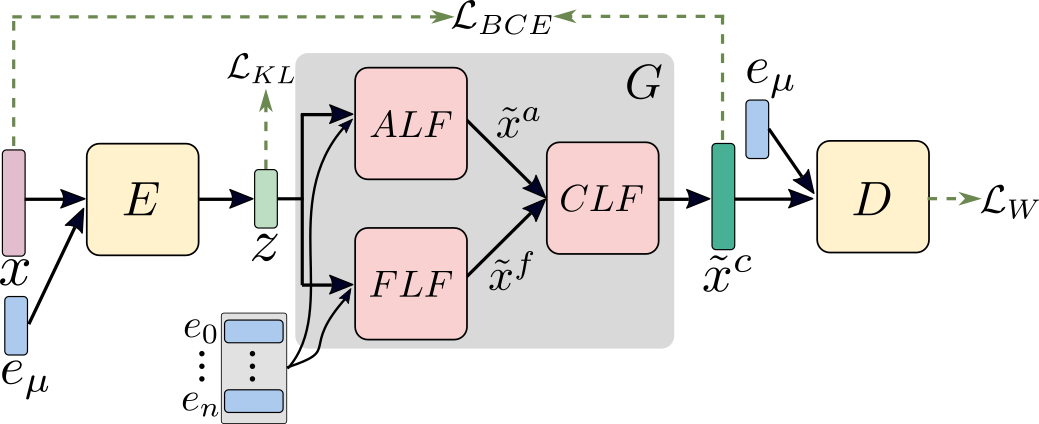}
    \caption{Our CLF-based \vaegan{} architecture, which integrates the proposed multi-label feature synthesis CLF approach into the \vaegan{}. The standard \vaegan{} extends \clswgan{} by integrating a conditional VAE along with a conditional WGAN, utilizing a shared generator between them. \vaegan{} comprises an encoder $E$, shared generator and a discriminator $D$, conditioned on the class-specific embeddings $e$. In the proposed CLF-based \vaegan{} shown here, we replace the single-label shared generator in \vaegan{} with our multi-label generator $G$ for synthesizing multi-label $\tilde{x}^{c}$. The resulting synthesized features $\tilde{x}^{c}$ are then passed to the discriminator for training the generator using the loss terms: $\mathcal{L}_{KL}$, $\mathcal{L}_{BCE}$ and $\mathcal{L}_{W}$.\vspace{-0.3cm}}
    \label{fig:overall_arch_vaegan}
\end{figure}

% f-vaegan
In addition to \clswgan{}, we also integrate our CLF into \vaegan{}~\cite{xian2019f} framework to perform multi-label feature synthesis. The \vaegan{} extends \clswgan{} by combining a conditional VAE~\cite{kingma13iclr} along with a conditional WGAN, utilizing a shared generator between them. For more details on \vaegan{}, we refer to~\cite{xian2019f}. Similar to our CLF-based \clswgan{} described earlier, we replace the single-label shared generator in \vaegan{} with our multi-label generator (CLF) for synthesizing multi-label $\tilde{x}^{c}$. The resulting CLF-based \vaegan{} is trained similar to the standard \vaegan{}, using the original loss formulation~\cite{xian2019f}. Fig.\ref{fig:overall_arch_vaegan} shows our CLF-based \vaegan{} architecture for multi-label (generalized) zero-shot classification.

\section{Experiments}

% \subsection{Experimental Setup\label{sec:exp_setup}}
\noindent\textbf{Datasets:} We evaluate our approach on three benchmarks: NUS-WIDE~\cite{nuswide}, Open Images~\cite{openimages} and MS COCO~\cite{coco}.
%for the classification task.
% 
The \textbf{NUS-WIDE} dataset comprises nearly $270$K images with $81$ human-annotated categories, in addition to the $925$ labels obtained from Flickr user tags. As in~\cite{huynh2020shared,zhang2016fast}, the $925$ and $81$ labels are used as seen and unseen classes, respectively. 
The \textbf{Open Images} (v4) is a large-scale dataset with $9$ million training images along with $41{,}620$ validation and $125{,}456$ testing images. 
% The scale of Open Images is larger than other multi-label datasets, such as NUS-WIDE and MS COCO. 
It is partially annotated with human and machine-generated labels. Here, $7{,}186$ labels, with at least $100$ training images, are selected as seen classes. The most frequent $400$ test labels, which are not present in the training data are selected as unseen classes, as in \cite{huynh2020shared}. 
% The \textbf{Open Images} (v4) is a large-scale dataset consisting of $9$ million training images along with $41{,}620$ and $125{,}456$ validation and testing images, respectively. The scale of Open Images is larger than other multi-label datasets, such as NUS-WIDE and MS COCO. This dataset is partially annotated with human labels and machine-generated labels. Here, $7{,}186$ labels, having at least $100$ images in the training, are selected as seen classes. The most frequent $400$ test labels, which are not present in the training data are selected as unseen classes, as in \cite{huynh2020shared}. 
% 
The \textbf{MS COCO} dataset is divided into training and validation sets with $82{,}783$ and $40{,}504$ images. Here, we perform multi-label zero-shot experiments by using the same split ($65$ seen and $15$ unseen classes), as in~\cite{bansal2018zero,hayat2020synthesizing}.\\
% The \textbf{MS COCO} dataset is divided into training and validation sets with $82{,}783$ and $40{,}504$ images. It has been previously used for multi-label zero-shot object detection~\cite{bansal2018zero,hayat2020synthesizing} but not for multi-label zero-shot classification. Here, we perform multi-label zero-shot classification experiments by using the same split ($65$ seen and $15$ unseen classes), as in detection works~\cite{bansal2018zero,hayat2020synthesizing}.\\
% 
%As in~\cite{bansal2018zero,hayat2020synthesizing}, the $80$ object categories are split into $65$ seen and $15$ unseen classes.
% 
% 
\noindent\textbf{Evaluation Metrics:}
Similar to~\cite{huynh2020shared,veit2017learning}, we evaluate our zero-shot classification approach using the F1 score at \textit{top}-$K$ predictions and the mean Average Precision (mAP). While the F1 captures the model's ability to correctly rank the labels in each image, the mAP metric captures the image ranking accuracy of the model for each label.\\
% For evaluating our approach on multi-label (generalized) zero-shot classification, we use the mean Average Precision (mAP) and F1 score at \textit{top}-$K$ predictions, similar to~\cite{huynh2020shared,veit2017learning,zhang2016fast}. While the mAP measure captures the image ranking accuracy of the model for each label, the F1 measure captures the model's ability to correctly rank the labels in each image.\\
% 
% For computing mAP, first the AP of each class is computed as $AP_j = \nicefrac{1}{N_j}\sum_{k=1}^{N}P(k,j)\cdot I(k,j)$, where 
% $P(k, j)$ is the precision for label $j$ at 
% $k$ best predictions, $I(k, j)$ is $1$ iff the label $j$ is in the ground-truth of
% the image at rank $k$ and $N_j$ is the number of images containing label $j$. Then, the mAP is computed as the mean of $AP_j$ over all labels $j$. To compute the F1 score, $top$-$K$ predicted classes are assigned to each image and mean precision and recall are computed as $P{=}\sum_j N_j^t/\sum_j N_j^p$ and $R{=}\sum_j N_j^t/\sum_j N_j$, respectively. Here, $N_j^p$ and $N_j^t$ denote the number of positive predictions and number of true positives for label $j$. Then, F1 score is computed as the harmonic mean between $P$ and $R$.
% 
\noindent\textbf{Multi-label Combinations:} Synthesizing multi-label features for learning the final multi-label ZSL and GZSL classifiers requires multi-label combinations with only unseen classes and also seen-unseen classes. 
To obtain these combinations, first, given a multi-label combination of seen classes from the training set, we randomly change the seen classes to their corresponding nearest unseen classes (based on distance between the class embeddings in the word-embedding space $\mathcal{E}$). This allows us to obtain different multi-label combinations involving only unseen classes and seen-unseen classes.  Consequently, our approach aids in the obtaining multi-label combinations that are more likely to be realistic and occurring in the test distribution.\\
\noindent\textbf{Implementation Details:}
Following existing zero-shot classification works~\cite{zhang2016fast,huynh2020shared}, we use the pretrained VGG-19 backbone to extract features from multi-label images.
%Multi-label images are input to a pretrained VGG-19~\cite{simonyan2014very} backbone for extracting visual features, as in~\cite{zhang2016fast,huynh2020shared}. 
Image-level features of size $d{=}4{,}096$ from FC$7$ layer output are used as input to our GAN. We use the $l_2$ normalized $300$ dimensional GloVe~\cite{pennington2014glove} vectors corresponding to the category names as the embeddings $e(k)\in\mathcal{E}$, as in~\cite{huynh2020shared}. The encoder $E$, discriminator $D$ and generators $G^a$ and $G^f$ are two layer fully connected (FC) networks with $4{,}096$ hidden units and Leaky ReLU non-linearity. The number of heads $H$ in our CLF is set to $8$. The sub-network $f(\cdot)$ in CLF is a two layer FC network with $8{,}192$ hidden units. 
The feature synthesizing network is trained with a $10^{-4}$ learning rate. The WGAN is trained with (\textit{batch size}, \textit{epoch}) of $(64,50)$, $(128,1)$ and $(64,70)$ on NUS-WIDE, Open Images and MS COCO, respectively. For the \clswgan{} variant, $\alpha$ is set to $0.1$, while $\lambda$ is set to $10$ in the \vaegan{} variant. The ZSL and GZSL classifiers: $f_{z}$ and $f_{gz}$ are trained for $50$ epochs with (\textit{batch size}, \textit{learning rate}) of $(300,10^{-3})$, $(100,10^{-4})$ and $(100,10^{-3})$ on NUS-WIDE, Open Images and MS COCO, respectively. We use the ADAM optimizer with ($\beta_1,\beta_2$) as ($0.5$, $0.999$) for training.
% The WGAN and the classifiers are trained using ADAM with ($\beta_1,\beta_2$) as ($0.5$, $0.999$). 
All the parameters are chosen via cross-validation. 

\begin{table}
\centering
\caption{Classification performance comparison of the three fusion approaches (ALF, FLF and CLF) for both ZSL and GZSL tasks on the NUS-WIDE dataset. The comparison is shown in terms of F1 score ($K{\in} \{3,5\}$) and mAP. We present the evaluation of our fusion approaches with both architectures: \clswgan{} and \vaegan{}. Regardless of the underlying generative architecture, our CLF approach achieves consistent improvement in performance, on all metrics, over both ALF and FLF for ZSL and GZSL tasks. Best results are in bold.\vspace{0.2em}}
\adjustbox{width=1\columnwidth}{
\begin{tabular}{cccccc} 
\toprule[0.15em]
\textbf{Fusion} & \textbf{GAN} & \textbf{Task} & \begin{tabular}[c]{@{}c@{}}\cellcolor[HTML]{EEEEEE} \textbf{~F1 }\textbf{(K=3)}\end{tabular} & \begin{tabular}[c]{@{}c@{}}\cellcolor[HTML]{DAE8FC} ~\textbf{F1}~\textbf{(K=5)}\end{tabular} & \textbf{mAP} \\ 
\toprule[0.15em]
\multirow{4}{*}{ALF} & \multirow{2}{*}{\texttt{f-CLSWGAN}} & ZSL & 29.8~ & 26.7 & 22.1 \\
 &  & GZSL & 16.8 & 19.7 & 7.1 \\
 & \multirow{2}{*}{\texttt{f-VAEGAN}} & ZSL & 30.9 & 27.9 & 22.9 \\
 &  & GZSL & 17.7 & 20.6 & 7.5 \\ 
\cmidrule(r){2-6}
% \midrule
\multirow{4}{*}{FLF} & \multirow{2}{*}{\texttt{f-CLSWGAN}} & ZSL & 29.8 & 26.8 & 22.6 \\
 &  & GZSL & 17.0 & 19.6 & 7.2 \\
 & \multirow{2}{*}{\texttt{f-VAEGAN}} & ZSL & 29.9 & 27.0 & 23.3 \\
 &  & GZSL & 17.2 & 19.9 & 7.6 \\ 
 \cmidrule(r){2-6}
% \midrule
\multirow{4}{*}{CLF} & \multirow{2}{*}{\texttt{f-CLSWGAN}} & ZSL & 31.1 & 27.6 & 23.7 \\
 &  & GZSL & 17.9 & 20.7 & 8.0 \\
 & \multirow{2}{*}{\texttt{f-VAEGAN}} & ZSL & \textbf{32.8} & \textbf{29.3} & \textbf{25.7} \\
 &  & GZSL & \textbf{18.9} & \textbf{22.0} & \textbf{8.9} \\
\bottomrule[0.1em]
\end{tabular}
}
\vspace{-0.2cm}
\label{tab:ablation}
\end{table}

\begin{table*}[t]
\centering
\caption{State-of-the-art comparison for ZSL and GZSL tasks on the NUS-WIDE and Open Images datasets. For a fair comparison, we use the same dataset splits (seen/unseen classes) as in LESA~\cite{huynh2020shared}. We report the results in terms of mAP and F1 score at $K{\in}\{3,5\}$ for NUS-WIDE and $K{\in}\{10,20\}$ for Open Images. Our approach outperforms the state-of-the-art for both ZSL and GZSL tasks, in terms of mAP and F1 score, on both datasets. Best results are in bold.}
\adjustbox{width=\linewidth}{
\begin{tabular}{ccccccccc|ccccccc} 
\toprule[0.15em]
\multirow{3}{*}{\textbf{ Method}} & \multirow{3}{*}{\textbf{Task }} & \multicolumn{7}{c}{\textbf{NUS-WIDE ( \#seen / \#unseen = 925/81) }} & \multicolumn{7}{c}{\textbf{Open Images ( \#seen / \#unseen = 7186/400) }} \\
 &  & \multicolumn{3}{c}{\cellcolor[HTML]{EEEEEE}\textbf{K = 3 }} & \multicolumn{3}{c}{\cellcolor[HTML]{DAE8FC}\textbf{K = 5 }} & \multirow{2}{*}{\textbf{mAP }} & \multicolumn{3}{c}{\cellcolor[HTML]{EEEEEE}\textbf{K = 10 }} & \multicolumn{3}{c}{\cellcolor[HTML]{DAE8FC}\textbf{K = 20 }} & \multirow{2}{*}{\textbf{mAP }} \\
 &  & \textbf{P } & \textbf{R } & \textbf{F1 } & \textbf{P } & \textbf{R } & \textbf{F1 } &  & \textbf{P } & \textbf{R } & \textbf{F1 } & \textbf{P } & \textbf{R } & \textbf{F1 } &  \\ 
\toprule[0.15em]
\multirow{2}{*}{CONSE~\cite{norouzi2013zero}} & ZSL & 17.5 & 28.0 & 21.6 & 13.9 & 37.0 & 20.2 & 9.4 & 0.2 & 7.3 & 0.4 & 0.2 & 11.3 & 0.3 & 40.4 \\
 & GZSL & 11.5 & 5.1 & 7.0 & 9.6 & 7.1 & 8.1 & 2.1 & 2.4 & 2.8 & 2.6 & 1.7 & 3.9 & 2.4 & 66.1 \\ 
\cmidrule(lr){2-16}
\multirow{2}{*}{LabelEM~\cite{akata2015label}} & ZSL & 15.6 & 25.0 & 19.2 & 13.4 & 35.7 & 19.5 & 7.1 & 0.2 & 8.7 & 0.5 & 0.2 & 15.8 & 0.4 & 40.5 \\
 & GZSL & 15.5 & 6.8 & 9.5 & 13.4 & 9.8 & 11.3 & 2.2 & 4.8 & 5.6 & 5.2 & 3.7 & 8.5 & 5.1 & 68.7 \\ 
\cmidrule(lr){2-16}
\multirow{2}{*}{Fast0Tag~\cite{zhang2016fast}} & ZSL & 22.6 & 36.2 & 27.8 & 18.2 & 48.4 & 26.4 & 15.1 & 0.3 & 12.6 & 0.7 & 0.3 & 21.3 & 0.6 & 41.2 \\
 & GZSL & 18.8 & 8.3 & 11.5 & 15.9 & 11.7 & 13.5 & 3.7 & 14.8 & 17.3 & 16.0 & 9.3 & 21.5 & 12.9 & 68.6 \\ 
\cmidrule(lr){2-16}
\multirow{2}{*}{One Attention per Label~\cite{kim2018bilinear}} & ZS & 20.9 & 33.5 & 25.8 & 16.2 & 43.2 & 23.6 & 10.4 & - & - & - & - & - & - & - \\
 & GZSL & 17.9 & 7.9 & 10.9 & 15.6 & 11.5 & 13.2 & 3.7 & - & - & - & - & - & - & - \\ 
\cmidrule(lr){2-16}
\multirow{2}{*}{One Attention per Cluster (M=10)~\cite{huynh2020shared}} & ZSL & 20.0 & 31.9 & 24.6 & 15.7 & 41.9 & 22.9 & 12.9 & 0.6 & 22.9 & 1.2 & 0.4 & 32.4 & 0.9 & 40.7 \\
 & GZSL & 10.4 & 4.6 & 6.4 & 9.1 & 6.7 & 7.7 & 2.6 & 15.7 & 18.3 & 16.9 & 9.6 & 22.4 & 13.5 & 68.2 \\ 
\cmidrule(lr){2-16}
\multirow{2}{*}{LESA (M=10)~\cite{huynh2020shared}} & ZSL & 25.7 & 41.1 & 31.6 & 19.7 & 52.5 & 28.7 & 19.4 & 0.7 & 25.6 & 1.4 & 0.5 & 37.4 & 1.0 & 41.7 \\
 & GZSL & 23.6 & 10.4 & 14.4 & 19.8 & 14.6 & 16.8 & 5.6 & 16.2 & 18.9 & 17.4 & 10.2 & 23.9 & 14.3 & 69.0 \\ 
\cmidrule(lr){2-16}
\multirow{2}{*}{\textbf{Our Approach}} & ZSL & \textbf{26.6}  & \textbf{42.8}  & \textbf{32.8}  & \textbf{20.1}  & \textbf{53.6}  & \textbf{29.3}  & \textbf{25.7}  & \textbf{1.3} & \textbf{42.4} & \textbf{2.5} & \textbf{1.1}  & \textbf{52.1} & \textbf{2.2} & \textbf{43.0} \\
 & GZSL & \textbf{30.9}  & \textbf{13.6}  & \textbf{18.9}  & \textbf{26.0}  & \textbf{19.1}  & \textbf{22.0}  & \textbf{8.9}  & \textbf{33.6} & \textbf{38.9} & \textbf{36.1} & \textbf{22.8} & \textbf{52.8} & \textbf{31.9} & \textbf{75.5} \\
\bottomrule[0.1em]
\end{tabular}
}
\vspace{-0.12cm}
\label{tab:sota_nuswide_openimages}
\end{table*}

\subsection{Ablation Study\label{sec:ablation}}
% \noindent\textbf{Generalization capabilities}:
We first present an ablation study w.r.t our fusion approaches: attribute-level (ALF), feature-level (FLF) and cross-level feature (CLF) on the NUS-WIDE dataset. We evaluate our fusion strategies with both architectures: \clswgan{} and \vaegan{}. Tab.~\ref{tab:ablation} shows the comparison, in terms of F1 score ($K{\in}\{3,5\}$) and mAP for both ZSL and GZSL tasks. In the case of ZSL, the ALF-based \clswgan{} achieves mAP score of $22.1$. The FLF-based \clswgan{} obtains similar performance with mAP score of $22.6$. The CLF-based \clswgan{} achieves improved performance with mAP score of $23.7$. Similarly, the CLF-based \clswgan{} obtains consistent improvement over both ALF and FLF-based \clswgan{} in terms of F1 score ($K{\in}\{3,5\}$). In the case of GZSL, ALF and FLF obtain  F1 scores ($K{=}3$) of $16.8$ and $17.0$, respectively. Our CLF approach achieves improved performance with F1 score of $17.9$. Similarly, CLF performs favorably against both ALF and FLF in terms of F1 at $K{=}5$ and mAP metrics.  

\begin{figure}
    \centering
    \includegraphics[width=0.98\columnwidth]{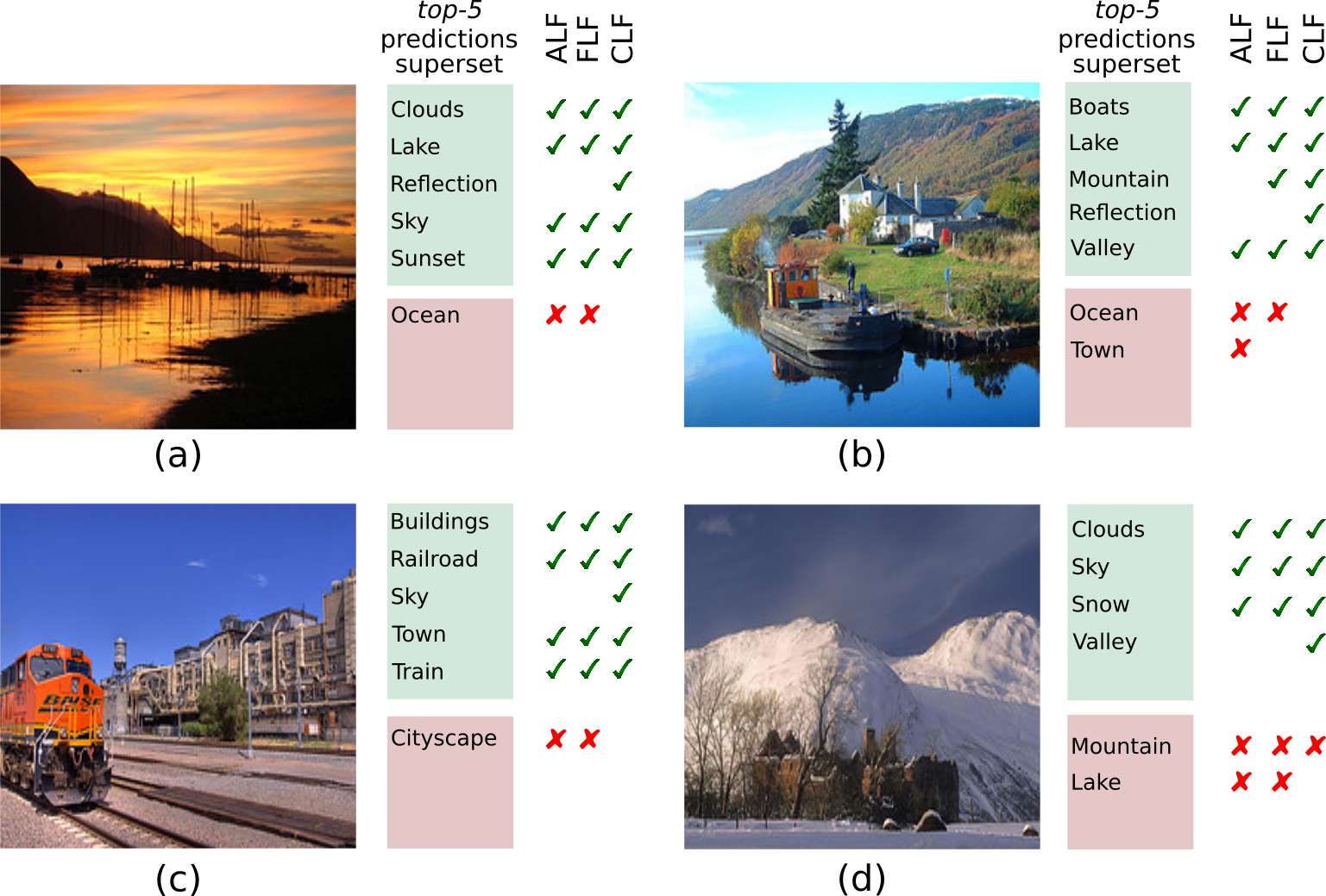}
    \caption{Qualitative results for the ZSL task on examples from NUS-WIDE. Alongside each example is a superset of \textit{top}-$5$ predictions from our ALF, FLF and CLF. The true- and false-positive classes are enclosed in green and red boxes. For each fusion approach, a green tick ({\color{ForestGreen}$\checkmark$}) and a red cross ({\color{red}\xmark}) is shown for true- and false-positive labels in its \textit{top}-$5$ predictions, respectively. Labels absent in the \textit{top}-$5$ predictions of a fusion have no {\color{ForestGreen}$\checkmark$} or {\color{red}\xmark}.
    % Generally, our ALF and FLF predict the true-positives reasonably well. 
    Our CLF, which integrates ALF and FLF, provides superior classification performance by correctly predicting classes missed by ALF and FLF, \textit{e.g.}, \textit{Reflection} in (a) and \textit{Sky} in (c), while removing false predictions such as, \textit{Ocean} in (b) and \textit{Lake} in (d).\vspace{-0.2cm}}
    \label{fig:clf_zsl_example}
\end{figure}

As in \clswgan{}, we also observe the CLF approach to achieve consistent improvement in performance over both ALF and FLF, when integrated in the more sophisticated \vaegan{}. In the case of ZSL, our CLF-based \vaegan{} achieves absolute gains of $2.8\%$ and $2.4\%$ in terms of mAP over ALF and FLF, respectively. Similarly, CLF-based \vaegan{} achieves consistent improvement in performance in terms of F1 score, over both ALF and FLF. Furthermore, CLF-based \vaegan{}  performs favorably against the other two fusion approaches in case of GZSL. Fig.~\ref{fig:clf_zsl_example} shows example qualitative results demonstrating the favorable performance of CLF over ALF and FLF. Additional qualitative results are illustrated in Sec.~\ref{sec:qual_res}.

%\highlight{The qualitative results in Fig.~\ref{fig:clf_zsl_example} show that CLF performs favorably in comparison to ALF and FLF. Additional results are provided in the supplementary.}

The aforementioned results show that our CLF approach achieves consistent improvement in performance over both ALF and FLF for both ZSL and GZSL, regardless of the underlying architecture. Moreover, the best results are obtained when integrating our CLF in \vaegan{}. 
Unless stated otherwise, we refer to our CLF-based \vaegan{} as ``Our Approach'' from here on.

\noindent\textbf{Varying the Multi-label Combinations for Feature Synthesis:} Here, we conduct an experiment to analyze the effect of generating unrealistic images from  completely random multi-label combinations and report the results in Tab.~\ref{tab:random_combination}. We observe that using random combinations decreases the multi-label classification performance in comparison to our approach of replacing a seen class with its nearest unseen class. Specifically, for the ZSL task, our approach obtains gains of 4.8, 3.2 and 4.7 in terms of F1 ($K$=3), F1 ($K$=5) and mAP, respectively, in comparison to employing random multi-label features. We also observe a similar trend in the case of GZSL task.

\begin{table}[t]
\centering
\caption{(G)ZSL performance comparison, on NUS-WIDE, between using (i)  random combinations of classes for synthesizing multi-label features and (ii) our approach of obtaining seen-unseen combinations from the train set multi-label combinations by replacing random seen classes  with their nearest unseen classes. We observe that learning the final ZSL and GZSL classifiers with features of multi-label combinations obtained by our approach significantly improves the zero-shot classification performance in comparison to learning with random multi-label combinations, which are likely to synthesize unrealistic features.\vspace{-0.2cm}}
\adjustbox{width=0.7\linewidth}{
\begin{tabular}{ccccc} 
% \toprule[0.15em]
\toprule[0.1em]

 \textbf{Method}  & \textbf{Task} & \begin{tabular}[c]{@{}c@{}}\cellcolor[HTML]{EEEEEE} \textbf{~F1 }\textbf{(K=3)}\end{tabular} & \begin{tabular}[c]{@{}c@{}}\cellcolor[HTML]{DAE8FC} ~\textbf{F1}~\textbf{(K=5)}\end{tabular} & \textbf{mAP} \\ 
% \toprule[0.15em]
\toprule

\multirow{2}{*}{Random} & ZSL & 28.0 & 26.1 & 21.0 \\
& GZSL & 15.8  & 17.9 & 6.1 \\
\midrule
\multirow{2}{*}{Ours} & ZSL & 32.8 & 29.3 & 25.7\\
& GZSL & 18.9 & 22.0 & 8.9 \\

\bottomrule[0.1em]
% \bottomrule[0.1em]
\end{tabular}%
}
\label{tab:random_combination}
\end{table}

\noindent\textbf{Varying the Fusion:} Here, we compare our cross-level feature fusion (CLF) with two fusion baselines: (i) averaging the output of ALF and FLF; (ii) concatenating the output of ALF and FLF, followed by an MLP. The results tabulated in Tab.~\ref{tab:baseline_fusion_ablation} show that the our CLF approach achieves significant gains over both \texttt{AVG} and \texttt{CONCAT} on both ZSL and GZSL tasks.

\begin{table}
\centering
\caption{(G)ZSL performance comparison of our CLF approach with baseline fusion approaches such as average (\texttt{AVG}) and concatenation (\texttt{CONCAT}) on the NUS-WIDE dataset, in terms of F1 and mAP. We observe that our cross-level feature fusion (CLF) approach achieves significant gains over both the baselines on both (G)ZSL tasks. \vspace{0.2em}}
\adjustbox{width=0.85\columnwidth}{
\begin{tabular}{ccccc} 
\toprule[0.15em]
 \textbf{Fusion} & \textbf{Task} & \begin{tabular}[c]{@{}c@{}}\cellcolor[HTML]{EEEEEE} \textbf{~F1 }\textbf{(K=3)}\end{tabular} & \begin{tabular}[c]{@{}c@{}}\cellcolor[HTML]{DAE8FC} ~\textbf{F1}~\textbf{(K=5)}\end{tabular} & \textbf{mAP} \\ 
\toprule[0.15em]

  \multirow{2}{*}{\texttt{AVG}} & ZSL & 30.2 & 27.4 & 23.2 \\
   & GZSL & 17.4 & 20.1 & 7.5 \\
   \midrule
  \multirow{2}{*}{\texttt{CONCAT}} & ZSL & 30.5 & 27.8 & 23.6 \\
   & GZSL & 17.8 & 20.6 & 7.7 \\
   \midrule
  \multirow{2}{*}{\texttt{Ours: CLF}} & ZSL & \textbf{32.8} & \textbf{29.3} & \textbf{25.7} \\
   & GZSL & \textbf{18.9} & \textbf{22.0} & \textbf{8.9} \\
 
\bottomrule[0.1em]
\end{tabular}
}
\vspace{-0.2cm}
\label{tab:baseline_fusion_ablation}
\end{table}

\noindent\textbf{Varying the Backbone:} Here, we experiment with image features extracted from different backbones and report the results in Tab.~\ref{tab:nuswide_backbone}. We observe that using stronger backbones improves the classification performance for both ZSL and GZSL tasks. \emph{E.g.}, employing a ResNet-50 pretrained in a self-supervised manner (DINO~\cite{caron2021emerging}) improves the mAP by 1.4 and 1.5 for the ZSL and GZSL tasks. Furthermore, unless otherwise stated, we employ VGG as the backbone to enable a fair comparison with existing multi-label zero-shot approaches~\cite{zhang2016fast,huynh2020shared}.

\begin{table}[t]
\centering
\caption{(G)ZSL performance comparison of our CLF approach on the NUS-WIDE dataset, when employing different backbones. We observe that the performance of our CLF for both (G)ZSL tasks improves when using stronger backbones like DINO ResNet-50~\cite{caron2021emerging}. }
\adjustbox{width=0.85\linewidth}{
\begin{tabular}{ccccc} 
% \toprule[0.15em]
\toprule[0.1em]

 \textbf{Backbone}  & \textbf{Task} & \begin{tabular}[c]{@{}c@{}}\cellcolor[HTML]{EEEEEE} \textbf{~F1 }\textbf{(K=3)}\end{tabular} & \begin{tabular}[c]{@{}c@{}}\cellcolor[HTML]{DAE8FC} ~\textbf{F1}~\textbf{(K=5)}\end{tabular} & \textbf{mAP} \\ 
% \toprule[0.15em]
\toprule
% \multirow{4}{*}{LESA [14]} & \multirow{2}{*}{Places365 ResNet-50} & ZSL & 19.8 & 31.7 & 28.8 \\
%  &  & GZSL & 6.0 & 15.6 & 17.1  \\
%  & \multirow{2}{*}{DINO ResNet-50} & ZSL & 20.5 & 32.1 & 29.0 \\
%  &  & GZSL & 7.5 & 16.9 & 17.6  \\
%  \hline
   \multirow{2}{*}{VGG~\cite{simonyan2014very}} & ZSL & 32.8 & 29.3 & 25.7\\
   & GZSL & 18.9 & 22.0 & 8.9 \\
   \midrule
 \multirow{2}{*}{Places365 ResNet-50~\cite{zhou2017places}} & ZSL  & 33.0 & 29.9 & 26.0 \\
   & GZSL  & 19.2 & 22.4 & 9.3 \\
 \midrule
  \multirow{2}{*}{DINO ResNet-50~\cite{caron2021emerging}} & ZSL & 33.7 & 30.7 & 27.1\\
   & GZSL & 20.1 & 23.2 & 10.4 \\
 
\bottomrule[0.1em]
% \bottomrule[0.1em]
\end{tabular}%
}
\vspace{-0.2cm}
\label{tab:nuswide_backbone}
\end{table}

\subsection{State-of-the-art Comparison\label{sec:sota_compare}}

\noindent\textbf{NUS-WIDE:} 
Tab.~\ref{tab:sota_nuswide_openimages} shows the state-of-the-art comparison\footnote{We exclude~\cite{ou2020multi,ji2020deep} from quantitative comparison due to the differences in training setup and evaluation metric computation.} for zero-shot (ZSL) and generalized zero-shot (GZSL) classification. The results are reported in terms of mAP and F1 score at \textit{top}-$K$ predictions ($K {\in} \{3,5\}$). In addition, we also report the precision (P) and recall (R) for each F1 score.  For the ZSL task, Fast0Tag~\cite{zhang2016fast}, which finds principal directions in the word vector space for ranking the relevant tags ahead of irrelevant tags, achieves an mAP of $15.1$. The recently introduced LESA \cite{huynh2020shared}, utilizing shared multi-attention to predict all labels in an image, achieves improved results over Fast0Tag, with an mAP of $19.4$. Our approach sets a new state of the art, outperforming LESA with an absolute gain of $6.3\%$ in terms of mAP. Similarly, our approach obtains consistent improvement in performance over the state-of-the-art in terms of F1 score ($K{\in}\{3,5\}$).
For the GZSL task, LESA obtains improved classification results among existing methods with an mAP of $5.6$. Our approach achieves mAP score of $8.9$, outperforming LESA with an absolute gain of $3.3\%$. Similarly, our approach achieves consistent improvement in performance with absolute gains of $4.5\%$ and $5.2\%$ over LESA in terms of F1 score at $K{=}3$ and $K{=}5$, respectively. 

\begin{figure}[t]
\centering
\includegraphics[width=0.8\columnwidth]{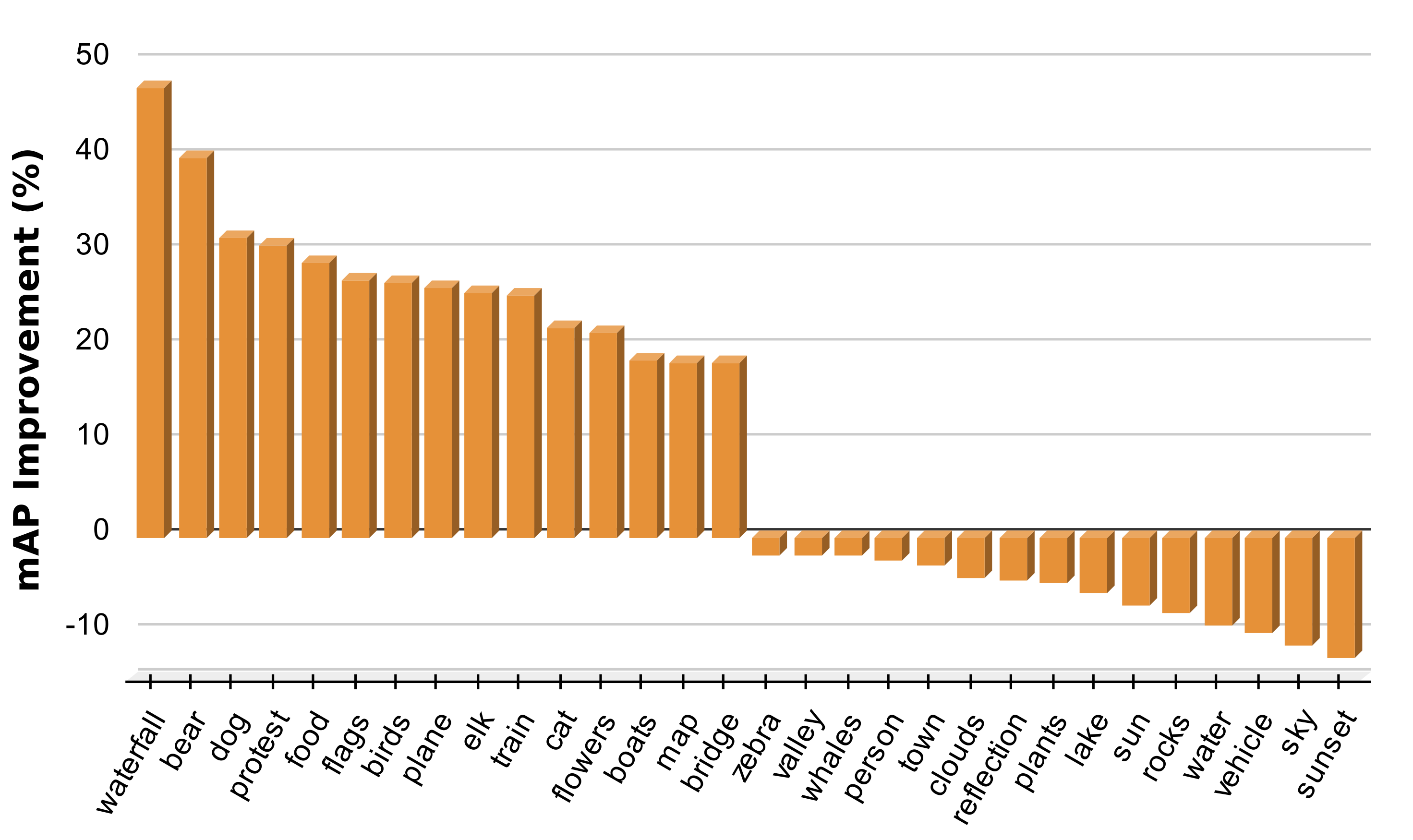}
\vspace{-0.2cm}
\caption{Classification mAP improvement comparison between our approach and LESA on NUS-WIDE. We show the comparison for $15$ unseen labels with the largest gain as well $15$ unseen labels with the largest drop. Our approach significantly improves (more than $26\%$) on several unseen labels (\textit{e.g.}, \textit{waterfall}, \textit{bear}, \textit{dog}, \textit{protest} and \textit{food}), while having relatively smaller (less than $12\%$) negative impact on other labels. Best viewed zoomed in.\vspace{-0.3cm}}
\label{fig:nuswide_compare}
\end{figure}

Fig.~\ref{fig:nuswide_compare} shows the classification comparison of our approach w.r.t LESA for $15$ unseen labels with the largest
performance gain as well as $15$ unseen labels with the largest drop. Our approach significantly improves (more than $26\%$) on several labels, such as \textit{waterfall}, \textit{bear}, \textit{dog}, \textit{protest} and \textit{food}, while having relatively smaller (less than $12\%$) negative impact on labels, such as \textit{sunset}, \textit{person}, \textit{sky}, \textit{water} and \textit{rocks}. We observe our approach to be particularly better on animal classes ($11$ out of $13$). In total, our approach outperforms LESA on $57$ out of $81$ labels. \\
% We observe our approach to be particularly better on animal classes ($11$ out of $13$), while struggling in some abstract concepts (\textit{e.g.} \textit{sunset}, \textit{sky}).
%On the large-scale Open Images dataset with 400 unseen categories, all methods.
%The results are reported in terms of mAP and F1 score at top K predictions ($K \in \{3,5\} $ for NUS-WIDE and $K \in \{10,20\} $ for Open-Images).
% 

% \begin{figure}[t]
% \centering
% \includegraphics[width=0.97\columnwidth]{images/openimages_zsl.png}
% \vspace{-0.2cm}
% \caption{}
% \label{fig:openimages_zsl_labelcount}
% \end{figure}

% 
%gives the performance comparison between our \proposed{} and state-of-the-art methods on the NUS-WIDE and Open Images datasets.
% 
% 
\begin{table}[t]
\centering
\caption{State-of-the-art comparison for ZSL and GZSL tasks on the MS COCO dataset having $65$ seen and $15$ unseen classes. We report the results in terms of mAP and F1 score at $K{=}3$ for ZSL and $K{=}\{3,5\}$ for GZSL. Our approach performs favorably against existing approaches for both ZSL and GZSL tasks. Best results are in bold.}\vspace{0.2em} 
\setlength{\tabcolsep}{12pt}
\adjustbox{width=1\linewidth}{
\begin{tabular}{ccccc} 
\toprule[0.15em]
\textbf{Method} & \textbf{Task} & \begin{tabular}[c]{@{}c@{}} \cellcolor[HTML]{EEEEEE}\textbf{F1 (K=3)} \end{tabular} & \begin{tabular}[c]{@{}c@{}}\cellcolor[HTML]{DAE8FC} \textbf{F1 (K=5)} \end{tabular} & \textbf{mAP} \\
\toprule[0.15em]
\multirow{2}{*}{CONSE~\cite{norouzi2013zero}} & ZSL & 18.4 & - & 13.2 \\
 & GZSL & 19.6 & 18.9 & 7.7 \\ 
\cmidrule(lr){2-5}
\multirow{2}{*}{LabelEM~\cite{akata2015label}} & ZSL & 10.3 & - & 9.6 \\
 & GZSL & 6.7 & 7.9 & 4.0 \\ 
\cmidrule(r){2-5}
\multirow{2}{*}{Fast0tag~\cite{zhang2016fast}} & ZSL & 37.5 & - & 43.3 \\
 & GZSL & 33.8 & 34.6 & 27.9 \\ 
\cmidrule(lr){2-5}
\multirow{2}{*}{LESA~\cite{huynh2020shared}} & ZSL & 33.6 & - & 31.8 \\
 & GZSL & 26.7 & 28.0 & 17.5 \\ 
\cmidrule(r){2-5}
% \multirow{2}{*}{\begin{tabular}[c]{@{}c@{}}\textbf{Ours:} With \\\clswgan~\end{tabular}} & ZS & 42.4 & - & 46.2 \\
%  & GZS & 38.3 & 39.4 & 30.3~ \\ 
% \cmidrule(lr){2-5}
\multirow{2}{*}{\textbf{Our Approach}} & ZSL & \textbf{43.5} & - & \textbf{52.2} \\
 & GZSL & \textbf{44.1} & \textbf{43.4} & \textbf{33.2} \\
\bottomrule[0.1em]
\end{tabular}%
}
% \vspace{-0.2cm}
\label{tab:sota_coco}
\end{table}

\noindent\textbf{Open Images:} Tab.~\ref{tab:sota_nuswide_openimages} shows the state-of-the-art comparison for ZSL and GZSL tasks. The results are reported in terms of mAP and F1 score at \textit{top}-$K$ predictions ($K {\in} \{10,20\}$). Since only $4{,}728$ seen classes are present in the testing set instances, the mAP score for the GZSL task is obtained %as the mean of 
by averaging AP for the $5{,}128$ (seen+unseen) classes, following~\cite{openimages}. Compared to NUS-WIDE, Open Images has significantly larger number of labels. This makes the ranking problem within an image more challenging, reflected by the lower F1 scores in the table. For the ZSL task, LESA obtains F1 scores of $1.4$ and $1.0$ at $K{=}10$ and $K{=}20$, respectively. Our approach performs favorably against LESA with F1 scores of $2.5$ and $2.2$ at $K {=} 10$ and $K {=} 20$, respectively. A similar performance gain is also observed for the mAP metric.
% Similarly, the proposed approach also achieves superior performance in terms of mAP. 
It is worth noting that this dataset has $400$ unseen labels, thereby making the problem of ZSL challenging. As in ZSL, our approach also achieves consistent gains, in both F1 and mAP, over the state-of-the-art for GZSL. 

% As in ZSL, our approach also achieves consistent improvement in performance, in terms of both F1 and mAP, over the state-of-the-art for the GZSL task. Although the problem of zero-shot object detection is investigated, 

\begin{table*}[t]
\centering
\caption{State-of-the-art comparison for ZSL and GZSL tasks on the NUS-WIDE and Open Images datasets \textit{using the proposed splits}. We report the results in terms of mAP and F1 score at $K{\in}\{3,5\}$ for NUS-WIDE. Similarly, F1 score for Open Images is computed at $K{\in}\{3,5\}$ for ZSL and $K{\in}\{10,20\}$ for GZSL. Our approach outperforms the state-of-the-art for both ZSL and GZSL tasks, in terms of mAP and F1 score, on both datasets. Best results are in bold. }
\setlength{\tabcolsep}{9pt}
\adjustbox{width=\linewidth}{
\begin{tabular}{ccccccccc|ccccccc} 
\toprule[0.15em]
\multirow{3}{*}{\textbf{ Method}} & \multirow{3}{*}{\textbf{Task }} & \multicolumn{7}{c}{\textbf{NUS-WIDE ( \#seen / \#unseen = 837/84) }} & \multicolumn{7}{c}{\textbf{Open Images ( \#seen / \#unseen = 7186/367) }} \\
 &  & \multicolumn{3}{c}{\cellcolor[HTML]{EEEEEE}\textbf{K = 3 }} & \multicolumn{3}{c}{\cellcolor[HTML]{DAE8FC}\textbf{K = 5 }} & \multirow{2}{*}{\textbf{mAP }} & \multicolumn{3}{c}{\cellcolor[HTML]{EEEEEE}\textbf{K = 3 (ZSL), 10 (GZSL) }} & \multicolumn{3}{c}{\cellcolor[HTML]{DAE8FC}\textbf{K = 5 (ZSL), 20 (GZSL) }} & \multirow{2}{*}{\textbf{mAP }} \\
 &  & \textbf{P } & \textbf{R } & \textbf{F1 } & \textbf{P } & \textbf{R } & \textbf{F1 } &  & \textbf{P } & \textbf{R } & \textbf{F1 } & \textbf{P } & \textbf{R } & \textbf{F1 } &  \\ 
\toprule[0.15em]
\multirow{2}{*}{Fast0Tag~\cite{zhang2016fast}} & ZSL & 20.1 & 23.3 & 21.6 & 17.4 & 33.6  & 22.9 & 13.9 & 2.6 & 3.1 & 2.8 & 2.9 & 9.6 & 4.5 & 62.1 \\
 & GZSL & 24.6 & 10.9 & 15.2 & 20.1 & 14.9 & 17.1 & 5.1 & 16.9 & 19.9 & 18.2 & 10.2 & 23.5 & 14.2 & 66.8 \\ 
\cmidrule(lr){2-16}
\multirow{2}{*}{LESA (M=10)~\cite{huynh2020shared}} & ZSL & 21.0 & 24.3 & 22.5 & 18.1 & 34.9 & 23.8 & 14.2 & 3.2 & 7.8 & 4.5 & 3.2 & 12.7 & 5.1 & 62.3 \\
 & GZSL & 29.0 & 12.9 & 17.8 & 23.6 & 17.5 & 20.1 & 6.5 & 18.3 & 21.1 & 19.6 & 11.2 & 25.8 & 15.6 & 67.2 \\
\cmidrule(lr){2-16}
\multirow{2}{*}{\textbf{Our Approach}}  & ZSL & \textbf{22.7} & \textbf{26.2} & \textbf{24.3} & \textbf{19.3} & \textbf{37.2} & \textbf{25.6} & \textbf{16.1} & \textbf{3.5} & \textbf{11.5} & \textbf{5.4} & \textbf{4.3} & \textbf{14.5} & \textbf{6.5} & \textbf{64.5} \\
 & GZSL & \textbf{30.2} & \textbf{13.5} & \textbf{18.6} & \textbf{25.2} & \textbf{18.8} & \textbf{21.4} & \textbf{9.5} & \textbf{33.6} & \textbf{38.8} & \textbf{36.0} & \textbf{22.7} & \textbf{52.8} & \textbf{31.8} & \textbf{75.3} \\
\bottomrule[0.1em]
\end{tabular}
}
% \vspace{-0.12cm}
\label{tab:new_split_sota_nuswide_openimages}
\end{table*}

\begin{table}[!t]
\centering
\caption{Impact of employing our CLF on top of recent discriminative approaches (LESA~\cite{huynh2020shared} and BiAM~\cite{narayan2021discriminative}). The performance of our CLF is also reported for ease of comparison. We observe that replacing the VGG backbone feature extractor in our CLF with LESA and BiAM significantly improves the performance in terms of F1 and mAP for (G)ZSL tasks over the respective discriminative approaches. These results show that our generative CLF approach can be easily integrated with state-of-the-art discriminative approaches to improve the multi-label zero-shot classification performance. \vspace{-0.2cm}}
\adjustbox{width=0.9\columnwidth}{
\begin{tabular}{ccccc} 
\toprule[0.15em]
\textbf{Method} & \textbf{Task} & \begin{tabular}[c]{@{}c@{}}\cellcolor[HTML]{EEEEEE} \textbf{~F1 }\textbf{(K=3)}\end{tabular} & \begin{tabular}[c]{@{}c@{}}\cellcolor[HTML]{DAE8FC} ~\textbf{F1}~\textbf{(K=5)}\end{tabular} & \textbf{mAP} \\ 
\toprule[0.15em]

 \multirow{2}{*}{\texttt{CLF}} & ZSL & 32.8 & 29.3 & 25.7 \\
  & GZSL & 18.9 & 22.0 & 8.9 \\

\midrule

\multirow{2}{*}{\texttt{LESA}~\cite{huynh2020shared}} & ZSL & 31.6 & 28.7 & 19.4 \\
& GZSL & 14.4 & 16.8 & 5.6  \\

\multirow{2}{*}{\texttt{LESA~\cite{huynh2020shared} + CLF}} & ZSL & 36.6 & 32.7 & 27.9 \\
 & GZSL & 20.4 & 23.1 & 8.2 \\
 \midrule
 \multirow{2}{*}{\texttt{BiAM}~\cite{narayan2021discriminative}} & ZSL & 33.1 & 30.7 & 26.3     \\
 & GZSL & 16.1 & 19.0 & 9.3 \\
 \multirow{2}{*}{\texttt{BiAM~\cite{narayan2021discriminative} + CLF}} & ZSL & \textbf{38.5} & \textbf{35.1} & \textbf{29.4} \\
 & GZSL & \textbf{23.2} & \textbf{25.3} & \textbf{10.2} \\

\bottomrule[0.1em]
\end{tabular}
}
\label{tab:clf_combined1}
\end{table}

\noindent\textbf{MS COCO:} We are the first to evaluate zero-shot classification on this dataset. Tab.~\ref{tab:sota_coco} shows the state-of-the-art comparison for ZSL and GZSL tasks. Since the maximum number of unseen classes per image is $3$ in \textit{val} set, we report the F1 at only $K{=}3$ for the ZSL task. We re-implement LabelEM, CONSE and Fast0tag since the respective source codes are not publicly available. For LESA, we obtain the results using the code-base provided by the authors.
%\highlight{For LESA, LabelEM and CONSE methods, we obtain the results by using the source code provided by the respective authors. For Fast0tag, we re-implement their approach since the code is not publicly available.}
Our approach obtains superior (G)ZSL performance, in both F1 score and mAP, compared to existing methods. Particularly, our approach obtains significant gains of $8.9\%$ and $5.3\%$ in terms of mAP for the ZSL and GZSL tasks over Fast0Tag~\cite{zhang2016fast}. Similar gains over the existing methods are also obtained for F1 score on both tasks, showing the efficacy of our approach.
%Our approach obtains superior (G)ZSL classification performance, in terms of F1 score and mAP, compared to existing methods. 

\begin{figure}[t]
    \centering
    \includegraphics[width=0.45\columnwidth]{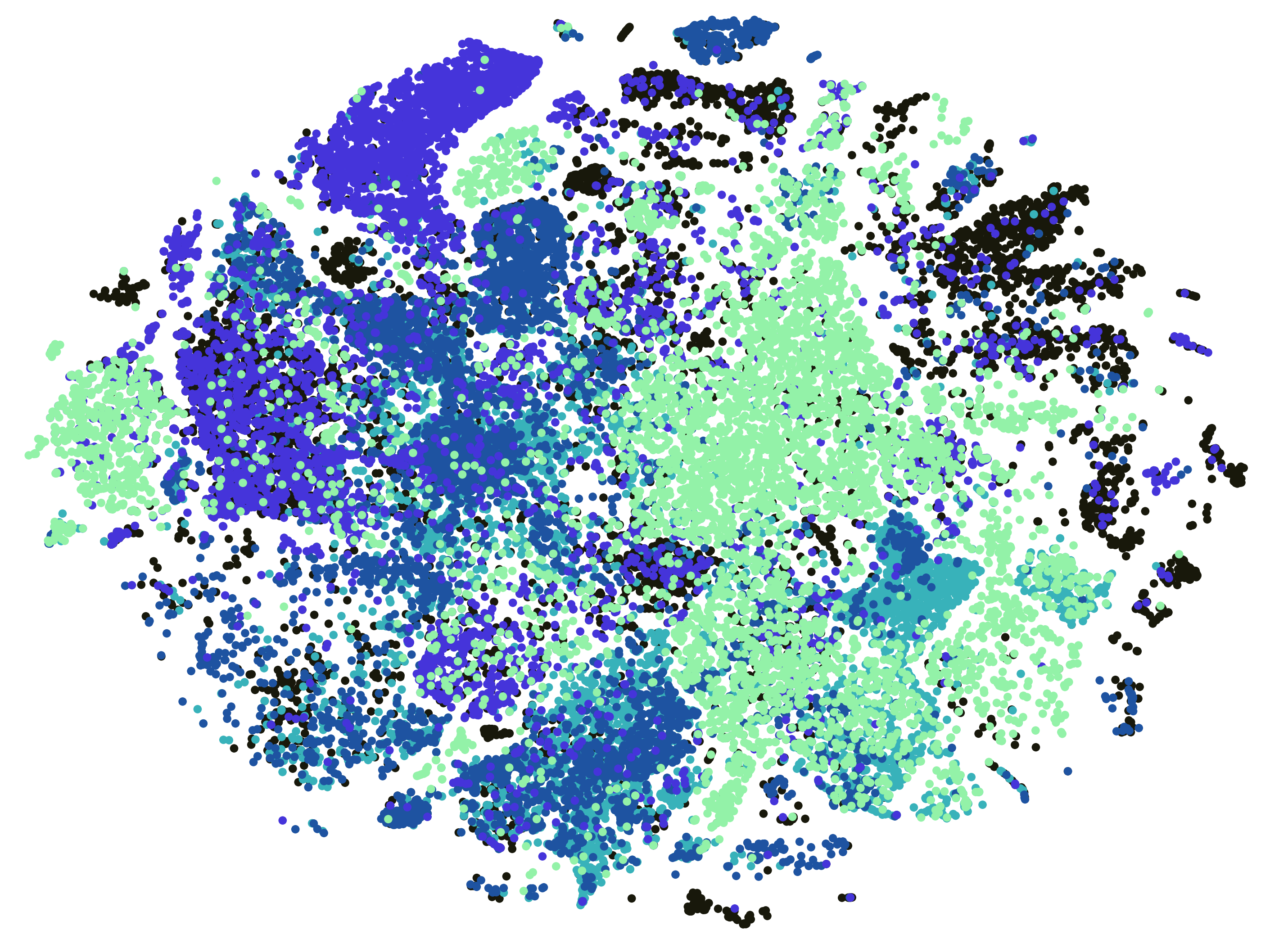}
    \includegraphics[width=0.45\columnwidth]{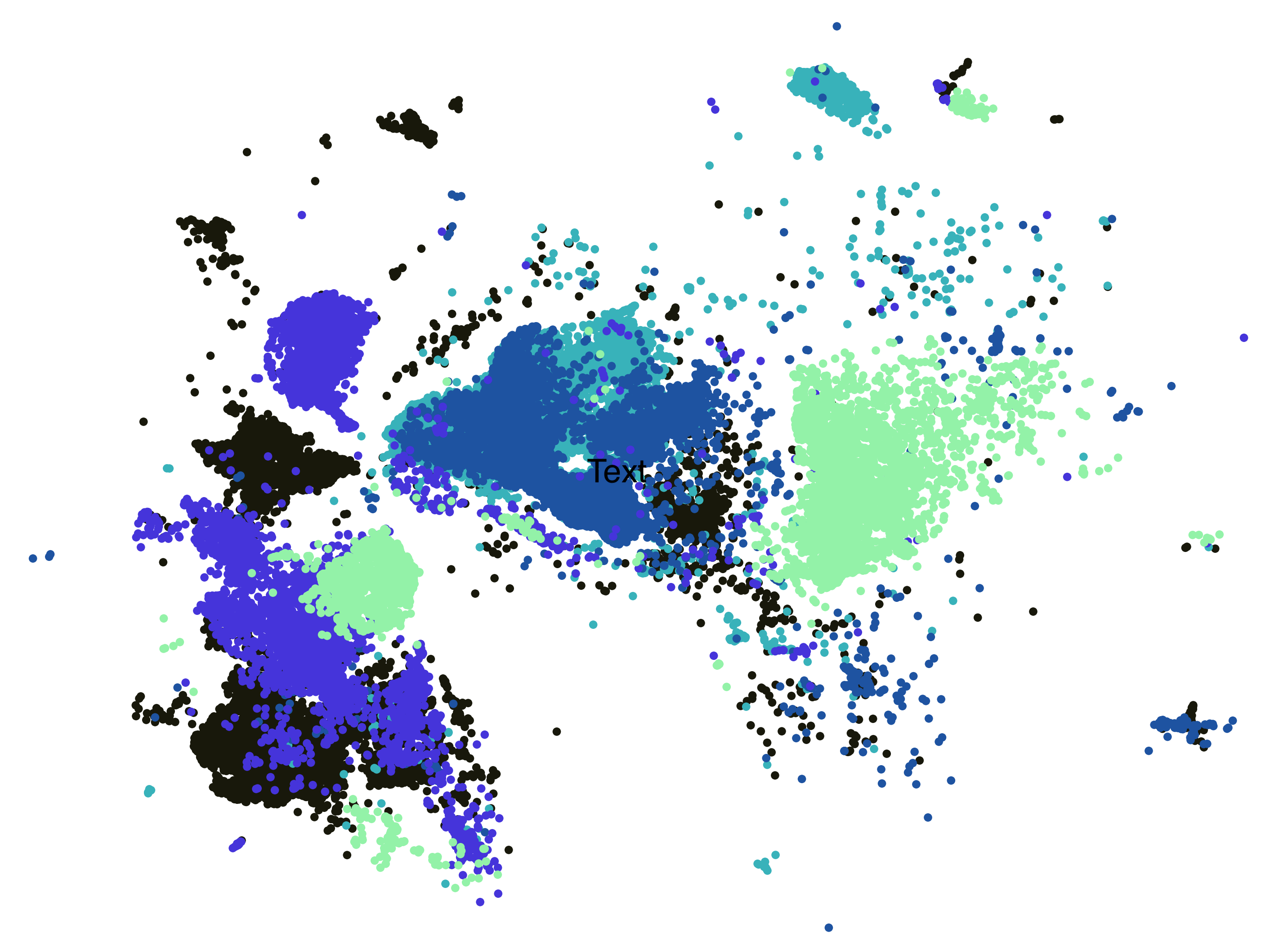}
    \caption{Qualitative comparison, in terms of t-SNE plots, between a recent discriminative approach BiAM~\cite{narayan2021discriminative} (on the left) and our CLF employed on top of BiAM (on the right). Visual feature instances from \textit{top}-$5$ frequently occurring unseen classes in the test set of the NUS-WIDE dataset are utilized for obtaining the t-SNE plots. The plots show that the features of different classes are better separated (and thereby discriminative) in the case of BiAM+CLF (on the right) compared to the class features from BiAM alone (on the left).}
    \label{fig:qual_tsne}
\end{figure}

\noindent\textbf{Impact of Employing CLF with Discriminative Approaches:} Here, we conduct an experiment for analyzing the effect of improved discriminability of the synthesized multi-label unseen class features. To this end, we employ the recent discriminative multi-label ZSL approaches LESA~\cite{huynh2020shared} and BiAM~\cite{narayan2021discriminative} as multi-label feature extractors and learn our CLF on these features. Tab.~\ref{tab:clf_combined1} shows that the ZSL and GZSL classification performance improves when employing our fusion-based CLF on top of these recent approaches. This is likely due to the discriminative approaches being trained on the seen class multi-label features alone and thereby relying on the mapping in the visual-semantic joint space for classifying novel unseen classes. However, in our CLF approach, zero-shot classifiers have better knowledge of the unseen classes since they are learned using the synthesized unseen class features. Furthermore, in Fig.~\ref{fig:qual_tsne}, t-SNE plots show the improvement in unseen class feature discriminability when CLF is learned on the BiAM features~\cite{narayan2021discriminative}.

% \begin{table}[t]
% \centering
% \caption{NUS-WIDE ablation with new split and with merged annotation}\vspace{0.2em} 
% \setlength{\tabcolsep}{12pt}
% \adjustbox{width=1\linewidth}{
% \begin{tabular}{ccccc} 
% \toprule[0.15em]
% \textbf{Method} & \textbf{Task} & \begin{tabular}[c]{@{}c@{}} \cellcolor[HTML]{EEEEEE}\textbf{F1 (K=3)} \end{tabular} & \begin{tabular}[c]{@{}c@{}}\cellcolor[HTML]{DAE8FC} \textbf{F1 (K=5)} \end{tabular} & \textbf{mAP} \\
% \toprule[0.15em]
% \multirow{2}{*}{Fast0tag~\cite{zhang2016fast}} & ZSL & 18.3 & 18.8 & 12.4 \\
%  & GZSL & 15.1 & 17.1 & 5.1 \\
% \cmidrule(lr){2-5}
% \multirow{2}{*}{LESA~\cite{huynh2020shared}} & ZSL & 18.2 & 20.4 & 12.1\\
%  & GZSL &  19.8 & 20.1 & 6.5 \\
% \cmidrule(r){2-5}
% \multirow{2}{*}{\textbf{Our Approach}} & ZSL & \textbf{} & \textbf{} & \textbf{} \\
%  & GZSL & \textbf{} & \textbf{} & \textbf{} \\
% \bottomrule[0.1em]
% \end{tabular}%
% }
% \vspace{-0.2cm}
% \label{tab:sota_coco}
% \end{table}

\subsection{State-of-the-art Comparison on Proposed Splits \label{sec:sota_new_splits}}

The seen/unseen class splits \cite{zhang2016fast,huynh2020shared} for NUS-WIDE and Open Images employed by existing approaches for evaluating on the multi-label (G)ZSL tasks do not strictly conform with the zero-shot paradigm. This is because the backbones used for feature extraction are pre-trained on the ImageNet~\cite{imagenet} dataset, whose classes overlap with a few unseen classes of NUS-WIDE and Open Images. Note that this issue does not arise for the MS COCO experiments since the backbone is retrained after removing the overlapping ImageNet classes, as in~\cite{hayat2020synthesizing}. Thus, in order to conform with the ZSL paradigm, we propose new seen/unseen splits for both NUS-WIDE and Open Images by ensuring that there is no overlap between the pre-trained ImageNet classes and the new unseen classes of the respective datasets. 

\noindent\textbf{NUS-WIDE:} We first preprocess the data to remove any label  inconsistencies. In the original split of $925$/$81$, there were multiple pairs with identical classes but minor spelling variations, \textit{e.g.}, \textit{harbor}-\textit{harbour}, \textit{window}-\textit{windows}, \textit{animal}-\textit{animals}. Such classes in each pair are merged and considered as a single class in the proposed split. Furthermore, classes originally in unseen set but overlapping with ImageNet classes are moved to the seen class set. Similarly, a few seen classes that had no overlap with ImageNet classes are moved to the unseen set to balance the splits. Consequently, the proposed (G)ZSL split for NUS-WIDE contains $837$ seen classes and $84$ unseen classes. \\
\noindent\textbf{Open Images:} The original $400$ unseen classes are present only in the test set and do not have any corresponding annotation in the $5.5$ million training images. Hence, only the unseen classes that overlap with the ImageNet classes are removed and not considered during evaluation. This results in $367$ unseen classes being retained in the proposed split. Furthermore, following~\cite{huynh2020shared}, testing instances without any unseen class annotations are not considered when evaluating for the ZSL task. In addition, since nearly $95\%$ and $99\%$ of the testing images have less than $3$ and $5$ unseen classes, respectively, we compute the F1 scores at $K {\in} \{3,5\}$ for the ZSL task. However, we retain $K {\in} \{10,20\}$ for GZSL, as in~\cite{huynh2020shared}. Further details regarding the classes in the proposed splits for both datasets are provided in the appendix.

Tab.~\ref{tab:new_split_sota_nuswide_openimages} shows the performance comparison of our method with existing approaches (Fast0Tag~\cite{zhang2016fast} and LESA~\cite{huynh2020shared}) on the proposed splits of NUS-WIDE and Open Images. Our approach performs favorably in comparison to existing methods, achieving significant gains over LESA up to $3\%$ and $8.1\%$ for NUS-WIDE and Open Images, in terms of mAP on the GZSL task. Similar gains for F1 scores are also achieved by our approach for both ZSL and GZSL. These results reinforce that \textit{irrespective of the seen/unseen class splits}, our approach performs favorably against existing methods on both ZSL and GZSL tasks in the multi-label setting.

\begin{table}[t]
\centering
\caption{Standard multi-label classification performance comparison on NUS-WIDE. The results are reported in terms of mAP and F1 score at $K{\in}\{3,5\}$. Our approach achieves superior performance compared to existing methods. Best results are in bold.}
\adjustbox{width=\linewidth}{
\begin{tabular}{cccccccc} 
\toprule[0.15em]
\multirow{2}{*}{\textbf{Method} } & \multicolumn{3}{c}{\cellcolor[HTML]{EEEEEE}\textbf{K=3}} & \multicolumn{3}{c}{\cellcolor[HTML]{DAE8FC}\textbf{K=5}} & \multirow{2}{*}{\textbf{mAP}} \\
 & \textbf{P} & \textbf{R} & \textbf{F1} & \textbf{P} & \textbf{R} & \textbf{F1} &  \\ 
\toprule[0.15em]
Logistic~\cite{tsoumakas2007multi} & 46.1 & 57.3 & 51.1 & 34.2 & 70.8 & 46.1 & 21.6 \\ 
\cmidrule(r){2-8}
WARP~\cite{gong2013deep} & 49.1 & 61.0 & 54.4 & 36.6 & 75.9 & 49.4 & 3.1 \\ 
\cmidrule(r){2-8}
WSABIE~\cite{weston2011wsabie} & 48.5 & 60.4 & 53.8 & 36.5 & 75.6 & 49.2 & 3.1 \\
\cmidrule(r){2-8}
Fast0Tag~\cite{zhang2016fast} & 48.6 & 60.4 & 53.8 & 36.0 & 74.6 & 48.6 & 22.4 \\
\cmidrule(r){2-8}
CNN-RNN~\cite{wang2016cnn} & 49.9 & 61.7 & 55.2 & 37.7 & 78.1 & 50.8 & 28.3 \\ 
\cmidrule(r){2-8}
One Attention per Label~\cite{kim2018bilinear} & 51.3 & 63.7 & 56.8 & 38.0 & 78.8 & 51.3 & 32.6 \\ 
\cmidrule(r){2-8}
One Attention per Cluster~\cite{huynh2020shared} & 51.1 & 63.5 & 56.6 & 37.6 & 77.9 & 50.7 & 31.7 \\ 
\cmidrule(r){2-8}
LESA~\cite{huynh2020shared} & 52.3 & 65.1 & 58.0 & 38.6 & 80.0 & 52.0 & 31.5 \\ 
\cmidrule(r){2-8}
\textbf{Our Approach} & \textbf{53.5} & \textbf{66.5} & \textbf{59.3} & \textbf{39.4} & \textbf{81.6} & \textbf{53.1} & \textbf{46.7} \\
\bottomrule[0.1em]
\end{tabular}%
}
\label{tab:mll_nuswide}
\end{table}

\subsection{Standard Multi-Label Classification\label{sec:stnd_ML_suppl}}
In addition to zero-shot classification, we evaluate our approach for standard multi-label classification where all the category labels during training and testing are identical. Tab.~\ref{tab:mll_nuswide} shows the state-of-the-art comparison on NUS-WIDE with $81$ human annotated labels. As in~\cite{huynh2020shared}, we remove all the testing instances that do not have any classes corresponding to the $81$ labels set. 
Among the existing methods, CNN-RNN and LESA achieve mAP scores of $28.3$ and $31.5$, respectively. Our approach outperforms existing methods by achieving an mAP score of $46.7$. Our method also performs favorably against these methods in terms of F1 scores.

Similarly, we also evaluate on the large-scale Open Images dataset~\cite{openimages}. Tab.~\ref{tab:mll_openimages} shows the state-of-the-art comparison for the standard multi-label classification on Open Images, with $7{,}186$ classes used for training and evaluating the approaches. As in~\cite{huynh2020shared}, test samples with missing labels for these $7{,}186$ classes are removed during evaluation. Furthermore, since only $4{,}728$ ground-truth classes are present in the test set, following~\cite{openimages}, the mAP score is obtained as the mean of AP for these classes only. Among the existing methods, Fast0Tag~\cite{zhang2016fast} and LESA~\cite{huynh2020shared} achieve an F1 score of $13.1$ and $14.5$ at $K{=}20$. Our approach outperforms the existing approaches by achieving an F1 score of $32.7$. The proposed approach also performs favorably in comparison to existing methods in terms of mAP score. Consequently, these performance gains over existing methods on both datasets for the standard multi-label classification task show the efficacy of our approach.

\begin{table}[t]
\centering
\caption{Standard multi-label classification performance comparison on Open Images. The results are reported in terms of mAP and F1 score at $K{\in}\{10,20\}$. Our approach achieves superior performance compared to existing methods. Best results are in bold.}
\adjustbox{width=\linewidth}{
\begin{tabular}{cccccccc} 
\toprule[0.15em]
\multirow{2}{*}{\textbf{Method} } & \multicolumn{3}{c}{\cellcolor[HTML]{EEEEEE}\textbf{K=10}} & \multicolumn{3}{c}{\cellcolor[HTML]{DAE8FC}\textbf{K=20}} & \multirow{2}{*}{\textbf{mAP}} \\
 & \textbf{P} & \textbf{R} & \textbf{F1} & \textbf{P} & \textbf{R} & \textbf{F1} &  \\ 
\toprule[0.15em]
Logistic~\cite{tsoumakas2007multi} & 12.1 & 14.7 & 13.3 & 8.4 & 20.2 & 11.8 & 75.1 \\   
\cmidrule(r){2-8}
WARP~\cite{gong2013deep} & 7.1 & 8.5 & 7.7 & 5.3 & 12.6 & 7.4 & 69.9 \\ 
\cmidrule(r){2-8}
WSABIE~\cite{weston2011wsabie} & 1.5 & 3.7 & 2.2 & 1.5 & 3.7 & 2.2 & 71.7 \\ 
\cmidrule(r){2-8}
Fast0Tag~\cite{zhang2016fast} & 14.9 & 17.9 & 16.2 & 9.3 & 22.3 & 13.1 & 69.0 \\ 
\cmidrule(r){2-8}
CNN-RNN~\cite{wang2016cnn} & 8.7 & 10.5 & 9.6 & 5.4 & 13.1 & 10.5 & 62.3 \\ 
% \cmidrule(r){2-8}
% One Attention per Label~\cite{kim2018bilinear} & - & - & - & - & - & - & - \\ 
\cmidrule(r){2-8}
One Attention per Cluster~\cite{huynh2020shared} & 14.9 & 17.9 & 16.3 & 9.2 & 22.0 & 13.0 & 68.5 \\
\cmidrule(r){2-8}
LESA~\cite{huynh2020shared} & 16.2 & 19.6 & 17.8 & 10.3 & 24.7 & 14.5 & 69.3 \\ 
\cmidrule(r){2-8}
\textbf{Our Approach} & \textbf{34.0} & \textbf{40.7} & \textbf{37.0} & \textbf{23.3} & \textbf{54.9} & \textbf{32.7} & \textbf{76.9} \\
\bottomrule[0.1em]
\end{tabular}%
}
\label{tab:mll_openimages}
\end{table}

%%% nus-wide mll

%In addition to zero-shot classification, we evaluate our approach for standard multi-label classification where all labels have training images. Tab.~\ref{tab:mll_nuswide} shows the state-of-the-art comparison for standard multi-label classification on NUS-WIDE with $81$ human annotated labels. As in~\cite{huynh2020shared}, we remove all test samples without any label in the $81$ label set. 
%Among existing methods, CNN-RNN and LESA achieve mAP scores of $28.3$ and $31.5$, respectively. Our approach outperforms existing methods, achieving mAP score of $46.7$. The proposed approach also performs favorably against existing methods in terms of F1 scores. 

% \section{Zero-shot Object Detection}
\section{Extension to Zero-shot Object Detection\label{sec:obj_det}}
Lastly, we also evaluate our multi-label feature generation approach (CLF) for zero-shot object detection (ZSD). ZSD strives for simultaneous classification and localization of previously unseen objects. In the generalized settings, the test set contains both seen and unseen object categories (GZSD). Recently, the work of~\cite{hayat2020synthesizing} introduce a zero-shot detection approach (SUZOD) where features are synthesized, conditioned upon class-embeddings, and integrated in the popular Faster R-CNN framework~\cite{ren2015faster}. The feature generation stage is jointly driven by the classification loss in the semantic embedding space for both seen and unseen categories. Their approach addresses multi-label zero-shot detection by constructing single-label features for each region of interest (RoI). Different from SUZOD~\cite{hayat2020synthesizing}, we first generate a pool of multi-label RoI features by integrating random sets of individual single-label RoI features. 
% These integrated multi-label RoI features are then used as real features to train our CLF-based classification architecture.  

To generate multi-label features for training our multi-label feature generation (CLF) approach, we first obtain the single-label features for foreground and background regions from the training images, as in SUZOD~\cite{hayat2020synthesizing}. These single-label features are taken from the output of \textit{ROI Align} layer in the Faster R-CNN framework~\cite{ren2015faster} and correspond to a single proposal obtained from the region proposal network (RPN). The foreground RoI-aligned seen class features of an image are then randomly combined to generate multiple multi-label features. These multi-label features are then passed through the two-layer fully connected network that follows the \textit{ROI Align} layer in the Faster R-CNN, to obtain $1024$-d multi-label features. The resulting seen class RoI features are then used as real features for training our multi-label feature generation (CLF) approach. Post-training of CLF, single-label unseen class features are synthesized and used for training the classifier for unseen classes. The learned classifier is then combined with the classifier head of the Faster R-CNN, as in SUZOD~\cite{hayat2020synthesizing}. Query images with unseen classes are then input to the modified Faster R-CNN framework to obtain detections for the unseen classes. 
% Additional details are provided in the supplementary.

\begin{figure}[t]
\begin{center}
% \adjustbox{width=0.98\columnwidth}{
\includegraphics[width=0.98\columnwidth]{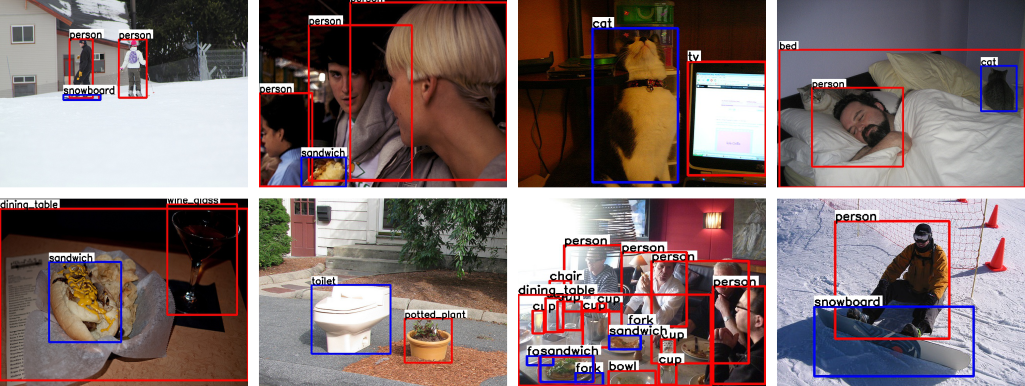}
% }
\vspace{-0.3cm}
\end{center}%\vspace{-1.5em}
\caption{\label{fig:detection}Qualitative detection results for GZSD on example images from MS COCO using our multi-label CLF-based detection approach. The seen and unseen class detections are shown in red and blue. Our detection approach achieves promising detection performance for both seen and unseen classes.\vspace{-0.25cm}}
\end{figure}

% \footnote{\label{footnote_additional_res}Additional results are provided in the supplementary material.}. 
Tab.~\ref{tab:sota_det_coco} shows the state-of-the-art comparison, in terms of mAP and recall, for ZSD and GZSD detection on MS COCO. For GZSD, Harmonic Mean (HM) of performances for seen and unseen classes are reported. Similar to SUZOD~\cite{hayat2020synthesizing}, we also report the results by using \clswgan{} as the underlying generative architecture. The polarity loss approach (PL-65)~\cite{rahman2018polarity} achieves an mAP of $12.4$ for the ZSD task, while the single-label feature generating approach of SUZOD~\cite{hayat2020synthesizing} achieves an improved performance of $19.0$ mAP. Our approach performs favorably against SUZOD with a significant gain of $1.3\%$ in terms of mAP.
Furthermore, while PL-65~\cite{rahman2018polarity} achieves a recall score of $37.72$ for ZSD, the SUZOD method~\cite{hayat2020synthesizing}
achieves a recall of $54.4$. Our CLF-based multi-label feature generative approach performs favorably in comparison to existing methods by achieving a recall score of $58.1$ for ZSD. Similarly, our approach also achieves consistent improvements for GZSD in terms of both mAP and recall. Fig.~\ref{fig:detection} shows qualitative detection results for GZSD. These results show that our approach improves the detection of unseen objects in (generalized) zero-shot settings.

\begin{table}[t]
\centering
\caption{State-of-the-art comparison for ZSD and GZSD tasks on MS COCO. The results are reported in terms of mean Average Precision (mAP) and Recall (Rec). For GZSD, we report the harmonic mean (HM) between seen and unseen classes. Our approach performs favorably against existing methods. Best results are in bold. }\vspace{0.2em}
\label{tab:sota_det_coco}
\adjustbox{width=\linewidth}{
\begin{tabular}{ccccccc} 
\toprule[0.15em]
\multirow{2}{*}{\textbf{Metric}} & \multirow{2}{*}{\textbf{Method}} & \multirow{2}{*}{\begin{tabular}[c]{@{}c@{}}~\textbf{Seen/Unseen} \\\textbf{Split}~ \end{tabular}} & \multirow{2}{*}{~\textbf{ZSD}~} & \multicolumn{3}{c}{\textbf{GZSD}} \\
 &  &  &  & ~\textbf{Seen}~ & ~\textbf{Unseen}~ & ~\textbf{HM}~ \\ 
\toprule[0.15em]
\multirow{6}{*}{{\textbf{mAP}}} & SB~\cite{bansal2018zero} & 48/17 & 0.70 & - & - & - \\
& DSES~\cite{bansal2018zero} & 48/17 & 0.54 & - & - & - \\
& PL-48~\cite{rahman2018polarity} & 48/17 & 10.01 & 35.92 & 4.12 & 7.39  \\
& PL-65~\cite{rahman2018polarity} & 65/15 & 12.40 & 34.07 & 12.40 & 18.18 \\
& SUZOD~\cite{hayat2020synthesizing} & 65/15 & 19.00 & 36.90 & 19.00 & 25.08 \\
& \textbf{Our Approach} & 65/15 & \textbf{20.30} & \textbf{37.60} & \textbf{20.30} & \textbf{26.36} \\
\cmidrule{2-7}
\multirow{6}{*}{{\textbf{Rec}}} & SB~\cite{bansal2018zero} & 48/17 & 24.39 & - & - & - \\
& DSES~\cite{bansal2018zero} & 48/17 & 27.19 & 15.02 & 15.32 & 15.17 \\
& PL-48~\cite{rahman2018polarity} & 48/17 & 43.56 & 38.24 & 26.32 & 31.18 \\
& PL-65~\cite{rahman2018polarity} & 65/15 & 37.72 & 36.38 & 37.16 & 36.76 \\
& SUZOD~\cite{hayat2020synthesizing} & 65/15 & 54.40 & 57.70 & 54.40 & 55.74 \\
& \textbf{Our Approach} & 65/15 & \textbf{58.10} & \textbf{58.90} & \textbf{58.10} & \textbf{58.50} \\

\bottomrule[0.1em]
\end{tabular}
}
% \vspace{-0.2cm}
\end{table}

\begin{figure*}[t]
    \centering
    \includegraphics[width=\textwidth]{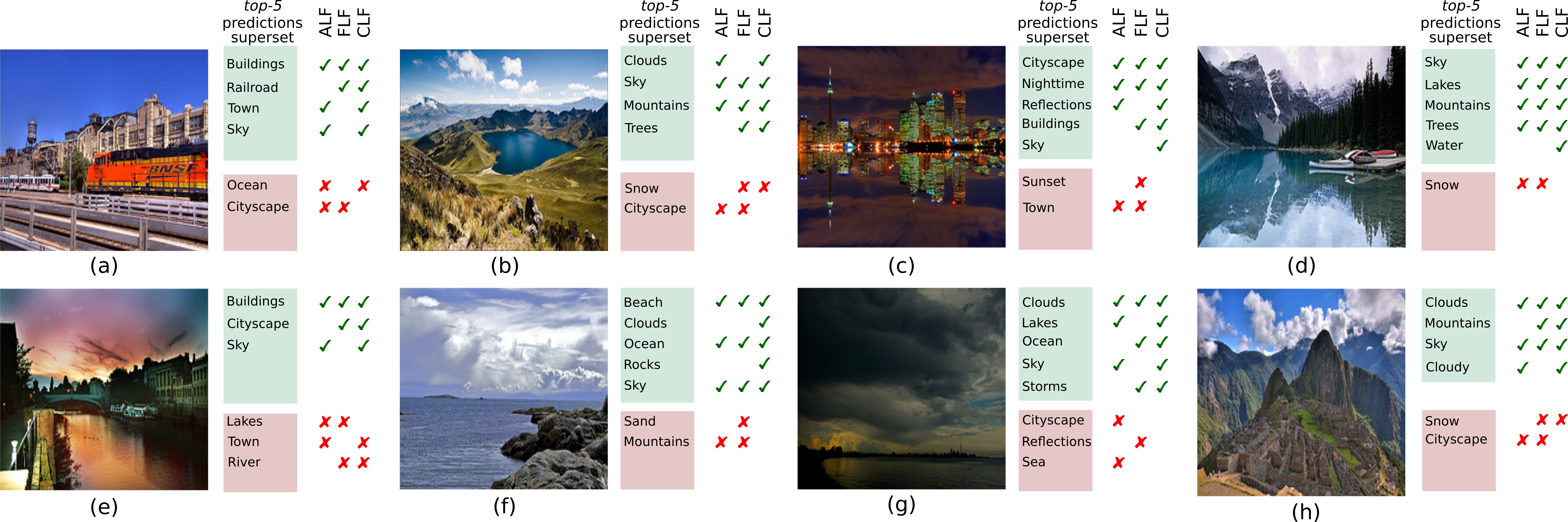}
    \caption{Qualitative results for multi-label ZSL on eight example images from the NUS-WIDE dataset. Alongside each example image is a superset of \textit{top}-$5$ predictions from our three fusion approaches: attribute-level (ALF), feature-level (FLF) and cross-level feature fusion (CLF). The true positive classes are enclosed in a green box. Similarly, false positives are enclosed in a red box. A green tick ({\color{ForestGreen}$\checkmark$}) is shown for a fusion approach if it has predicted the corresponding true positive label in its \textit{top}-$5$ predictions. Similarly, a red cross~({\color{red}\xmark}) is shown for a  false-positive label in its \textit{top}-$5$ predictions. A label without a tick or cross for a fusion approach denotes that its absence in the \textit{top}-$5$ predictions. Generally, our ALF and FLF approaches predict the true positives reasonably well. Our CLF, which integrates ALF and FLF approaches, improves the classification further with increased true positives and reduced false positives. Our CLF approach correctly predicts classes that were missed by both ALF and FLF, \textit{e.g.}, \textit{Sky} in image (c) and \textit{Water} in image (d). Furthermore, our CLF removes the false predictions of \textit{Town} and \textit{Snow} in (c) and (d), respectively. In addition, classes such as \textit{Mountains} in (h) and \textit{Reflections} in (c) are predicted correctly by either ALF or FLF, while both labels are correctly predicted by CLF. A similar observation holds in image (g). These results show that our CLF combines the advantages of both ALF and FLF, resulting in improved performance.}
    \label{fig:zsl_classification}
\end{figure*}

\begin{figure*}[t]
    \centering
    \includegraphics[width=\textwidth]{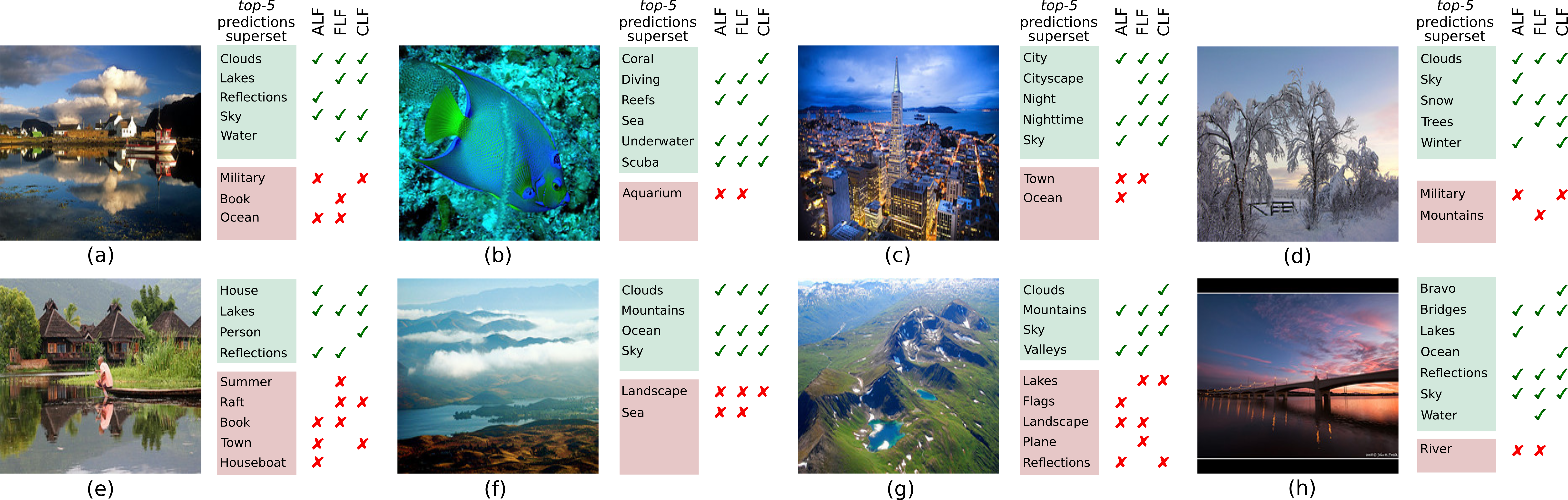}
    \caption{Qualitative results for multi-label GZSL on eight example images from the NUS-WIDE dataset. While our ALF and FLF approaches predict the true positives to a reasonable extent, our CLF improves the classification further with increased true positives and reduced false positives. Our CLF approach correctly predicts classes that were missed by both ALF and FLF, \textit{e.g.}, \textit{Coral}, \textit{Sea} in image (b) and \textit{Mountains} in image (f). False positives such as \textit{Ocean} and \textit{Aquarium} in (a) and (b) are also removed by our CLF. Furthermore, our CLF also correctly predicts classes such as \textit{Cityscape}, \textit{Sky} and \textit{Night} in image (c), which were predicted by either ALF or FLF. These results show that our CLF, which integrates our ALF and FLF approaches, achieves consistent improvement in performance over both ALF and FLF. See also Fig.~\ref{fig:zsl_classification} for more details.}
    \label{fig:gzsl_classification}
\end{figure*}

% \clearpage

\begin{figure*}[t]
\begin{center}
\scalebox{0.98}{
\begin{tabular}[t]{c} 
\includegraphics[width=\textwidth]{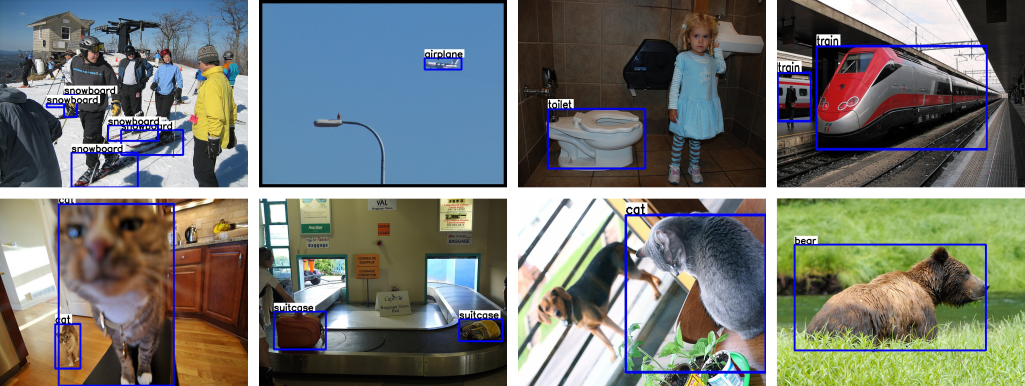} \\
(a) \textbf{Zero-shot object detection (ZSD)} \vspace{0.2cm} \\
\includegraphics[width=\textwidth]{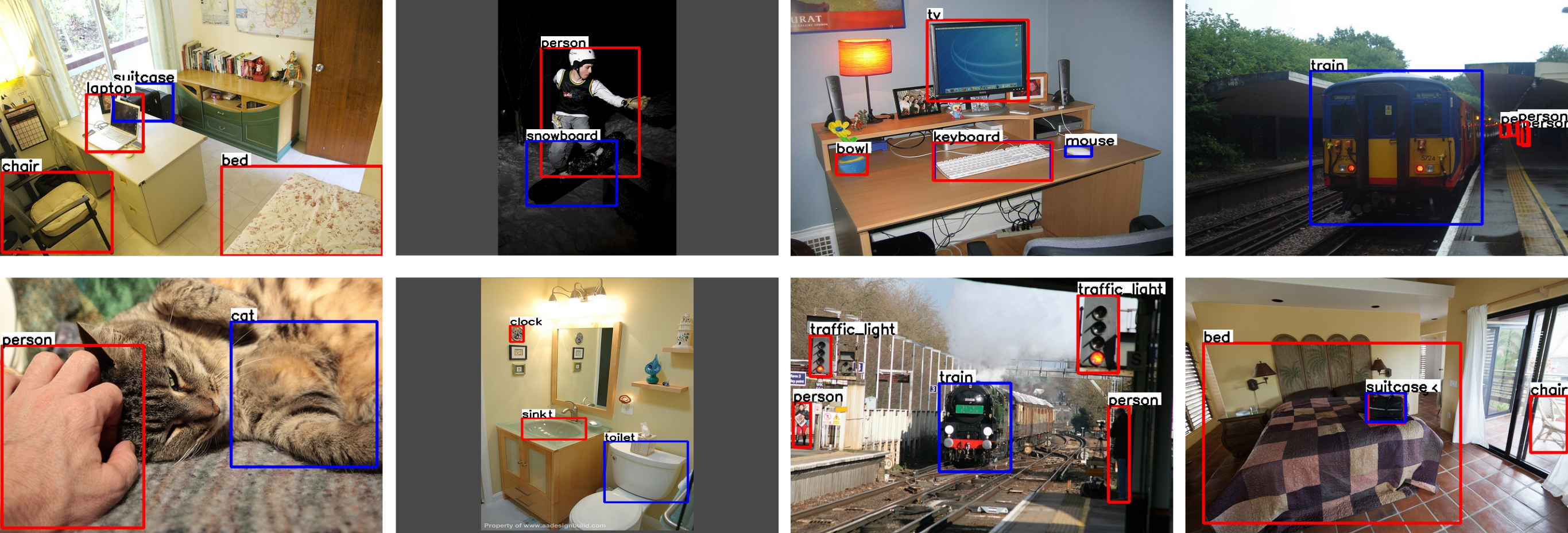} \\
(b) \textbf{Generalized zero-shot object detection (GZSD)}
\end{tabular}
}
\end{center}%\vspace{-1.5em}
\caption{\label{fig:detection_suppl}Qualitative results for zero-shot object detection on example images from the MS COCO dataset using our CLF approach. The zero-shot setting is illustrated in (a) followed by the generalized zero-shot setting in (b). The unseen classes detections are shown in blue, while the seen class predictions are in red. The unseen classes, such as \textit{snowboard}, \textit{airplane}, \textit{train} and \textit{cat} are correctly detected under zero-shot setting in (a), though the visual examples of these classes were not available for training. Similarly, we see that both unseen and seen classes are detected accurately under generalized zero-shot setting (b). These results show that our multi-label feature generation approach (CLF) achieves reasonably accurate object detection in (generalized) zero-shot settings.}
\label{fig:}
\end{figure*}

\section{Additional Qualitative Results\label{sec:qual_res}}
\noindent\textbf{Multi-Label Zero-Shot Classification:} Additional qualitative results are shown for multi-label zero-shot learning (ZSL) and generalized zero-shot learning (GZSL) in Fig.~\ref{fig:zsl_classification} and~\ref{fig:gzsl_classification}, respectively. Each figure comprises nine example images from the test set of the NUS-WIDE dataset~\cite{nuswide}. Alongside each example image is a list with the superset of \textit{top}-$5$ predictions from our three fusion approaches: attribute-level (ALF), feature-level (FLF) and cross-level feature fusion (CLF). The true positives, i.e, prediction labels matching the ground-truth classes in the image, are enclosed in a green box. Similarly, false positives, i.e, prediction labels that do not match with the ground-truth classes are enclosed in a red box. A green tick ({\color{ForestGreen}$\checkmark$}) is shown for a fusion approach if it has predicted the corresponding true positive label in its \textit{top}-$5$ predictions. Similarly, a red cross~({\color{red}\xmark}) is shown for a fusion approach if it has predicted the corresponding false-positive label in its \textit{top}-$5$ predictions. A label without a tick or cross for a fusion approach denotes its absence in the \textit{top}-$5$ predictions. 

In general, we observe that our ALF and FLF approaches predict true positives reasonably well in their \textit{top}-$5$ predictions. Furthermore, our CLF approach, which combines the advantages of both ALF and CLF, increases the true positives and reduces the false positives, leading to consistent performance gains over both ALF and FLF. Our CLF approach correctly predicts classes missed by both ALF and FLF, \textit{e.g.}, \textit{Sky} in Fig.~\ref{fig:zsl_classification}(c), \textit{Clouds} in Fig.~\ref{fig:zsl_classification}(f), \textit{Coral} in Fig.~\ref{fig:gzsl_classification}(b) and \textit{Mountains} in Fig.~\ref{fig:gzsl_classification}(f). Moreover, our CLF also removes false positives such as \textit{Snow} in Fig.~\ref{fig:zsl_classification}(d) and \textit{Town} in Fig.~\ref{fig:gzsl_classification}(c). In addition, our CLF correctly predicts classes such as \textit{Railroad} in Fig.~\ref{fig:zsl_classification}(a), \textit{Lakes} in Fig.~\ref{fig:zsl_classification}(g),  \textit{Cityscape}, \textit{Sky} and \textit{Night} in Fig.~\ref{fig:gzsl_classification}(c), which were predicted by either ALF or FLF. These results show the efficacy of our CLF approach that integrates our ALF and FLF, leading to improved (generalized) zero-shot classification.

\noindent\textbf{Zero-Shot Object Detection:} Fig.~\ref{fig:detection_suppl} shows additional qualitative results of our CLF approach on example images from the MS COCO dataset~\cite{coco} for the zero-shot object detection task. The zero-shot setting with only unseen classes detected is shown in Fig.~\ref{fig:detection_suppl}(a), followed by the generalized zero-shot setting in Fig.~\ref{fig:detection_suppl}(b), where both unseen and seen classes are detected. While the unseen classes detections are denoted by blue bounding boxes, the seen class detections are shown in red. We observe that the unseen classes, such as \textit{snowboard}, \textit{airplane}, \textit{train} and \textit{cat} are correctly detected under zero-shot setting in Fig.~\ref{fig:detection_suppl}(a), though the visual examples of these classes were not available for training. Furthermore, in the generalized zero-shot setting in Fig.~\ref{fig:detection_suppl}(b), we see that both unseen and seen classes are also detected accurately. These results show that our multi-label feature generation approach (CLF) achieves reasonably accurate object detection in (generalized) zero-shot settings.

In summary, our multi-label feature generation (CLF) approach, which integrates our attribute-level and feature-level fusion approaches, performs favorably in comparison to existing approaches on various tasks including (generalized) zero-shot multi-label classification, standard multi-label classification and (generalized) zero-shot detection. 

% Extension to Zero-shot Detection:

% We generate multi-label roi features by combining single-label roi features at a feature level by using the 7X7 feature map generated by rpn layer before passing it through shared 2 fc layers. We maintain the same fg:bg ratio as maintained by (SUZOD) paper. For seen and unseen classifier, we trained from scratch using the same multi-label seen roi features with a binary cross entropy loss. 

% Instead of using the faster-rcnn classifier head which is trained using a smooth L1-loss which is not suitable for our problem. We train a single fc layer seen classifier from scratch using the same multi-label roi fetaures.

% \appendices

% % you can choose not to have a title for an appendix
% % if you want by leaving the argument blank
% \section{}

% % use section* for acknowledgment
% \ifCLASSOPTIONcompsoc
%   % The Computer Society usually uses the plural form
%   \section*{Acknowledgments}
% \else
%   % regular IEEE prefers the singular form
% \fi

\section{Conclusion}
We investigate the problem of multi-label feature synthesis in the zero-shot setting. To this end, we introduce three fusion approaches: ALF, FLF and CLF. Our ALF synthesizes features by integrating class-specific attribute embeddings at the generator input.
% Our ALF produces an image-level embedding vector by integrating class-specific attribute embeddings at input of the generator. 
The FLF synthesizes features from class-specific embeddings individually and integrates them in feature space. Our CLF combines the advantages of both ALF and FLF, using each individual-level feature and attends to the bi-level context.
Consequently, individual-level features adapt themselves producing enriched synthesized features that are pooled to obtain the final output. 
% Consequently, individual-level features are enriched and pooled to obtain final output. 
We integrate our fusion approaches in two generative architectures. Our approach outperforms existing zero-shot methods on three benchmarks. Additionally, our approach performs favorably against existing methods for the standard multi-label classification task on two large-scale benchmarks. Lastly, we also show the effectiveness of our approach for zero-shot detection. We hope our simple and effective approach will serve as a solid baseline and help ease future research in generative multi-label zero-shot learning.

\section{Acknowledgment}
The authors would like to acknowledge the support from Grant PID2021-128178OB-I00.

% \ifCLASSOPTIONcaptionsoff
%   \newpage
% \fi



%\bibliographystyle{IEEEtran}

% H.~Kopka and P.~W. Daly, \emph{A Guide to {\LaTeX}}, 3rd~ed.\hskip 1em plus

% \end{thebibliography}

{
% \newpage
\small
\bibliographystyle{IEEEtran}
\bibliography{main}
}


% \newpage
% \begin{IEEEbiographynophoto}{Akshita Gupta}
\begin{IEEEbiography}[{\includegraphics[width=1in,height=1.2in,clip,keepaspectratio]{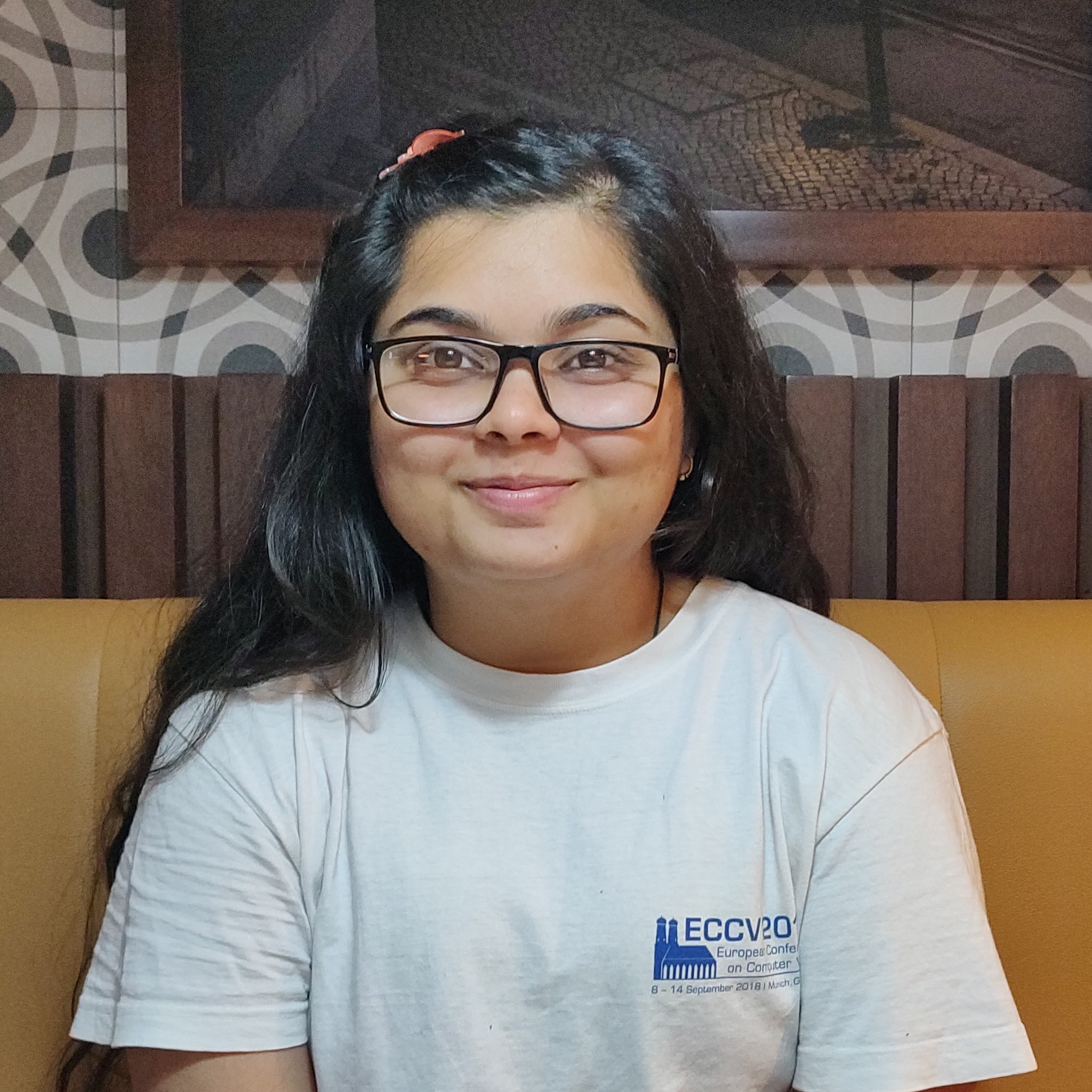}}]{Akshita Gupta}
is a MASc student at University of Guelph and Vector Institute for Artificial Intelligence. Previously, she was a Data Scientist at Bayanat LLC, Abu Dhabi and a  Research Engineer at the Inception Institute of Artificial Intelligence. She serves as a reviewer for CVPR, ECCV, ICCV, and TPAMI. She has worked as an Outreachy intern at Mozilla in 2018. Her research interest include low-shot (few-, zero-) classification, semantic/instance segmentation, object detection, and open-world learning.

% computer vision and machine learning.
% She completed her Bachelor degree from DIT University and, during that time, she studied a semester at Indian Institute of Technology, Roorkee. 
\end{IEEEbiography}

\vspace{-1.0cm}

\begin{IEEEbiography}[{\includegraphics[width=1in,height=1.2in,clip,keepaspectratio]{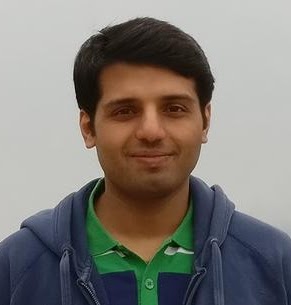}}]{Sanath Narayan}
% \begin{IEEEbiographynophoto}{Sanath Narayan}
is a Research Scientist at the Technology Innovation Institute, Abu Dhabi. He previously worked as a Research Scientist at Inception Institute of Artificial Intelligence and as a Senior Technical Lead at Mercedes-Benz R\&D India. He has served as a program committee member for several premier conferences including CVPR, ICCV and ECCV. He has been recognized as an outstanding/top reviewer multiple times at these conferences. He received his Ph.D. degree from the Indian Institute of Science in 2016. His thesis was awarded the best doctoral symposium paper award at ICVGIP 2014. His research interests include computer vision and machine learning.
\end{IEEEbiography}

\vspace{-1.0cm}

% \begin{IEEEbiographynophoto}{Salman Khan}
\begin{IEEEbiography}[{\includegraphics[width=1in,height=1.2in,clip,keepaspectratio]{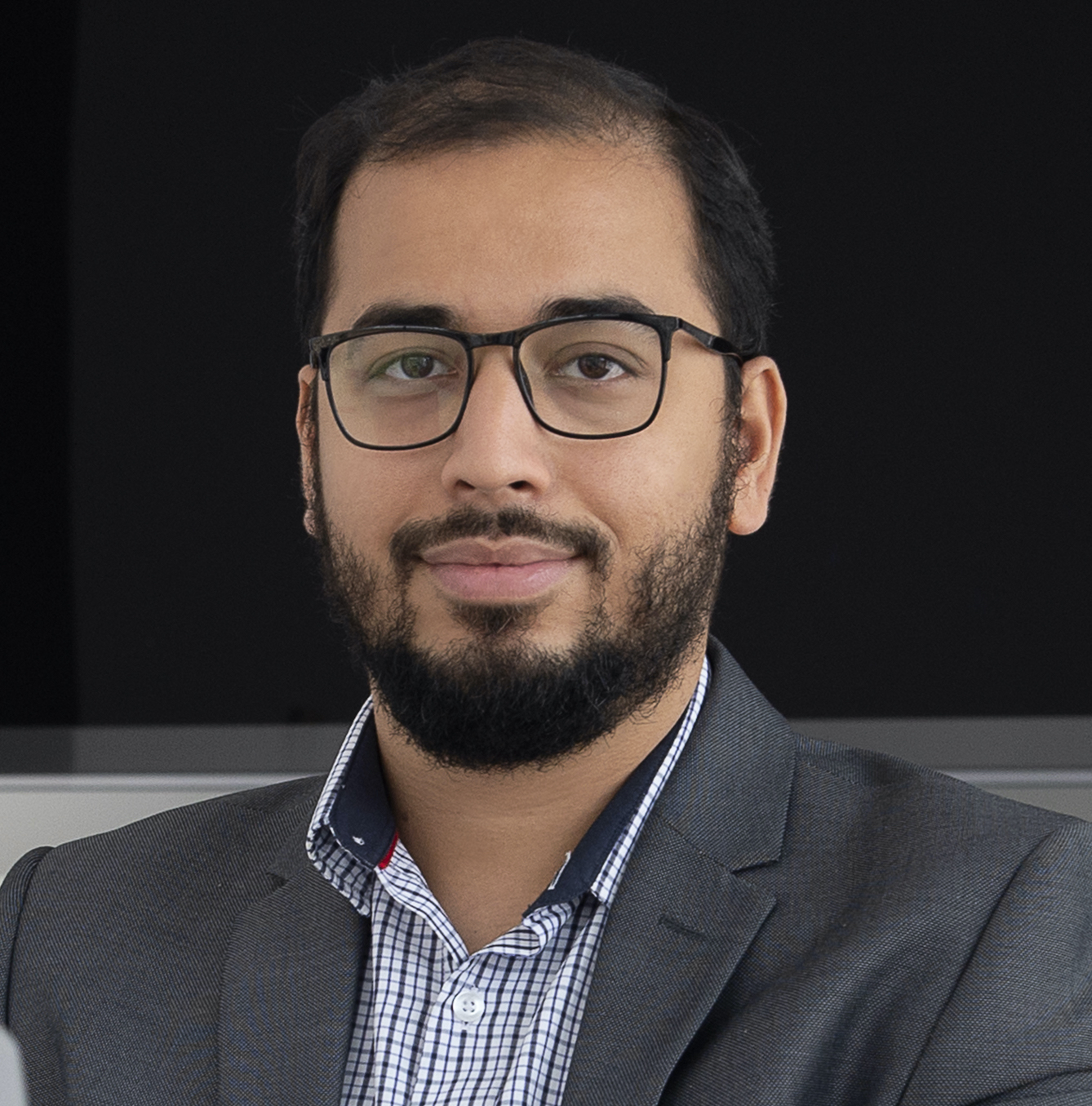}}]{Salman Khan}
is an Associate Professor at MBZ University of Artificial Intelligence. He has been an Adjunct faculty with Australian National University since 2016. 
%He worked as a Senior Scientist with the Inception Institute of AI (2018-2020) and as a Research Scientist with Data61-CSIRO (previously NICTA) from 2016-2018. 
He has been awarded the outstanding reviewer award at IEEE CVPR multiple times, won the best paper award at 9th ICPRAM 2020, and won 2nd prize in the NTIRE Image Enhancement Competition alongside CVPR 2019. He served as a (senior) program committee member for several premier conferences including CVPR, ICCV, ICML, ECCV and NeurIPS. He received his Ph.D. degree from The University of Western Australia in 2016. His thesis received an honorable mention on the Dean’s List Award. He has published over 100 papers in high-impact scientific journals and conferences. His research interests include computer vision and machine learning.
\end{IEEEbiography}

%\newpage

\vspace{-1.0cm}

% \begin{IEEEbiographynophoto}
% {Fahad Khan}
\begin{IEEEbiography}[{\includegraphics[width=1in,height=1.2in,clip,keepaspectratio]{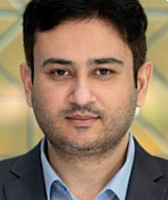}}]{Fahad Khan}
is currently a faculty member at MBZUAI, United Arab Emirates and Linkoping University, Sweden. 
%From 2018 to 2020 he worked as a Lead Scientist at the Inception Institute of Artificial Intelligence (IIAI), Abu Dhabi, United Arab Emirates. 
He received the M.Sc. degree in Intelligent Systems Design from Chalmers University of Technology, Sweden and a Ph.D. degree in Computer Vision from Autonomous University of
Barcelona, Spain. He has achieved top ranks on various international challenges (Visual Object Tracking VOT: 1st 2014 and 2018, 2nd 2015, 1st 2016; VOT-TIR: 1st 2015 and 2016; OpenCV Tracking: 1st 2015; 1st PASCAL VOC 2010). His research interests include a wide range of topics within computer vision and machine learning, such as object recognition, object detection, action recognition and visual tracking. He has published more than 100 articles in high-impact computer vision journals and conferences in these areas. He serves as a regular senior program committee member for leading computer vision conferences such as CVPR, ICCV, and ECCV.
\end{IEEEbiography}

\vspace{-1.0cm}

% \begin{IEEEbiographynophoto}{Ling shao}
\begin{IEEEbiography}[{\includegraphics[width=1in,height=1.2in,clip,keepaspectratio]{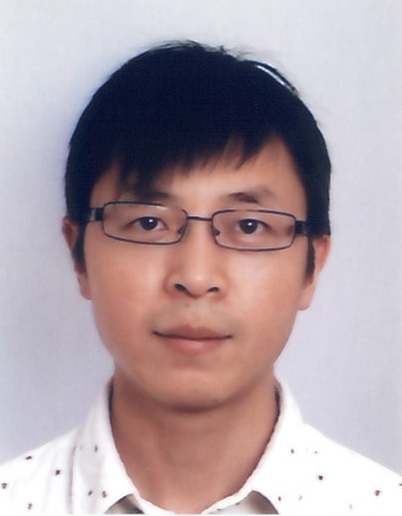}}]{Ling shao} 
is a Distinguished Professor with the UCAS-Terminus AI Lab, University of Chinese Academy of Sciences, Beijing, China. He was the founding CEO and Chief Scientist of the Inception Institute of Artificial Intelligence, Abu Dhabi, UAE. He was also the Initiator, founding Provost and EVP of MBZUAI. His research interests include generative AI, vision and language, and AI for healthcare. He is a fellow of the IEEE, the IAPR, the BCS and the IET.
\end{IEEEbiography}

\vspace{-1.0cm}

\begin{IEEEbiography}[{\includegraphics[width=1in,height=1.2in,clip,keepaspectratio]{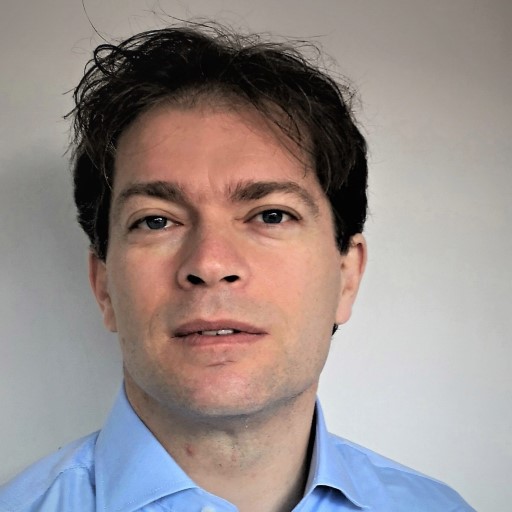}}]{Joost van de Weijer} 
% \begin{IEEEbiographynophoto}{Joost van de Weijer}
received the Ph.D. degree from the University of Amsterdam in 2005. He was a Marie Curie Intra-European Fellow at INRIA Rhone-Alpes, France, and from 2008 to 2012 was a Ramon y Cajal Fellow at the Universitat Autònoma de Barcelona, Spain, where he is currently a Senior Scientist at the Computer Vision Center and leader of the Learning and Machine Perception (LAMP) Team. His main research directions are color in computer vision, continual learning, active learning, and domain adaptation.
\end{IEEEbiography}


%\vfill


\appendix[Proposed ZSL Splits]
Here, we tabulate the unseen classes in the proposed ZSL splits of both datasets: NUS-WIDE~\cite{nuswide} and Open Images~\cite{openimages}. While the $84$ unseen classes in the new split of NUS-WIDE are given in Tab.~\ref{tab:nus_unseen}, the $367$ unseen classes of Open Images are given in Tab.~\ref{tab:openimages_unseen}.
Due to the large number of the seen classes in both datasets, they have been listed in our code repository.

\begin{table*}[t]
\caption{84 unseen classes in the proposed split of the NUS-WIDE dataset\vspace{-0.1cm}}
\label{tab:nus_unseen}
\centering
\begin{tabular}{@{}lllllll@{}}
\toprule
actor      & dance       & glacier     & mountains & protesters  & skyscraper & temple     \\
beach      & dancing     & grass       & museum    & railroad    & snow       & tower      \\
book       & diving      & harbour     & night     & rainbow     & statue     & town       \\
buildings  & dress       & helicopters & nighttime & reflections & storms     & trees      \\
canoe      & earthquake  & human       & ocean     & river       & streets    & village    \\
cellphones & fire        & hut         & outside   & road        & sun        & warehouse  \\
cemetery   & firefighter & lakes       & park      & rocks       & sunny      & water      \\
city       & flags       & landscapes  & people    & running     & sunset     & waterfalls \\
cityscape  & food        & leaves      & person    & sand        & sunshine   & wedding    \\
clouds     & frost       & maps        & police    & sea         & surf       & windows    \\
cloudy     & gardens     & microphones & pool      & sky         & swimmers   & winter     \\
coast      & girls       & moon        & protest   & skyline     & tattoo     & zoo        \\ \bottomrule
\end{tabular}
% \vspace{-0.3cm}
\end{table*}
\begin{table*}[!h]
\caption{367 unseen classes in the proposed split of the Open Images dataset\vspace{-0.1cm}}
\label{tab:openimages_unseen}
\centering
\tiny
\begin{tabular}{lllll}
\toprule
4 × 100 metres relay      & Abseiling                       & Academic certificate              & Acerola                                    & Agaricomycetes                     \\
Airbus a330               & Akita inu                       & Amaryllis belladonna              & American bulldog                           & American shorthair                 \\
Amphibious assault ship   & Amphibious transport dock       & Anchor handling tug supply vessel & Apiales                                    & Aquifoliaceae                      \\
Aquifoliales              & Araneus                         & Armored cruiser                   & Asian food                                 & Aubretia                           \\
Australian rules football & Australian silky terrier        & Auto race                         & Baker                                      & Balinese                           \\
Ballroom dance            & Bandy                           & Barque                            & Barquentine                                & Barramundi                         \\
Barrel racing             & Bass                            & Basset artésien normand           & Beach handball                             & Bell 412                           \\
Bengal                    & Bentley continental flying spur & Biceps curl                       & Biewer terrier                             & Big mac                            \\
Birman                    & Blackbird                       & Blt                               & Bmw 3 series (e90)                         & Bmw 320                            \\
Bmw 335                   & Bmw 5 series                    & Boa                               & Bodyguard                                  & Boeing                             \\
Boeing 717                & Boeing 787 dreamliner           & Boeing c-40 clipper               & Bombardier challenger 600                  & Bombycidae                         \\
Boston butt               & Bouillabaisse                   & Breckland thyme                   & Brig                                       & Brigantine                         \\
British bulldogs          & British semi-longhair           & British shorthair                 & Broholmer                                  & Bruschetta                         \\
Bullmastiff               & Burmese                         & Burmilla                          & Burnet rose                                & Caesar salad                       \\
Calochortus               & Camellia sasanqua               & Canaan dog                        & Canard                                     & Canoe slalom                       \\
Canoe sprint              & Caravel                         & Caridean shrimp                   & Carnitas                                   & Carom billiards                    \\
Carrack                   & Catalan sheepdog                & Caucasian shepherd dog            & Cavapoo                                    & Celesta                            \\
Cessna 150                & Cessna 172                      & Cessna 185                        & Cessna 206                                 & Champignon mushroom                \\
Char kway teow            & Charadriiformes                 & Chassis                           & Cheesesteak                                & Chicory                            \\
Chinese hawthorn          & Churchill tank                  & Cinclidae                         & Coccoloba uvifera                          & Cod                                \\
Colias                    & Colorado blue columbine         & Common rudd                       & Common snapping turtle                     & Condor                             \\
Conformation show         & Corn chowder                    & Crane vessel (floating)           & Cross-country skiing                       & Cuatro                             \\
Curtiss p-40 warhawk      & Custom car                      & Czechoslovakian wolfdog           & Dame’s rocket                              & Damson                             \\
Datura inoxia             & Day (Unit of time)              & Deep sea fish                     & Degu                                       & Dirt track racing                  \\
Diving                    & Dobermann                       & Douglas sbd dauntless             & Dragon                                     & Dredging                           \\
Drentse patrijshond       & Drever                          & Drilling rig                      & East siberian laika                        & Eastern diamondback rattlesnake    \\
Emberizidae               & Endurocross                     & English draughts                  & Escargot                                   & Estonian hound                     \\
Eurasier                  & European food                   & European garden spider            & European green lizard                      & Everlasting sweet pea              \\
Exercise                  & F1 Powerboat Racing             & Fast attack craft                 & Fauna                                      & Fawn lily                          \\
Finnish hound             & First generation ford mustang   & Flatland bmx                      & Floating production storage and offloading & Flora                              \\
Fluyt                     & Flxible new look bus            & Focke-wulf fw 190                 & Forage fish                                & Ford model a                       \\
Fortepiano                & Four o'clock flower             & Four o'clocks                     & Free solo climbing                         & Freediving                         \\
Freestyle swimming        & Fried prawn                     & Frittata                          & Fudge                                      & Galeas                             \\
Galgo español             & German pinscher                 & German spitz                      & German spitz klein                         & German spitz mittel                \\
Gladiolus                 & Go                              & Grumman f8f bearcat               & Gulfstream iii                             & Gulfstream v                       \\
Gumbo                     & Gumdrop                         & Gunboat                           & Gymnast                                    & Ham                                \\
Hapkido                   & Hawthorn                        & Heavy cruiser                     & Hula                                       & Hunting                            \\
Hydrangea serrata         & Hyssopus                        & Icelandic sheepdog                & Icon                                       & Jackup rig                         \\
Jaguar mark 1             & Jaguar xk150                    & Japanese Camellia                 & Japanese rhinoceros beetle                 & Jämthund                           \\
Karelian bear dog         & Keelboat                        & Khinkali                          & Kishu                                      & Kobe beef                          \\
Korat                     & Kuy teav                        & Lagotto romagnolo                 & Laksa                                      & Lamborghini                        \\
Lamborghini aventador     & Landslide                       & Large-flowered evening primrose   & Light cruiser                              & Lingonberry                        \\
Lissotriton               & Longship                        & Lovebird                          & Lugger                                     & Mackerel                           \\
Maglev                    & Maidenhair tree                 & Maine coon                        & Manchester terrier                         & Manila galleon                     \\
Manx                      & Matsutake                       & Mcdonnell douglas dc-9            & Mcdonnell douglas f-15e strike eagle       & Mcdonnell douglas f-4 phantom ii   \\
Media player              & Mercedes-benz e-class           & Microvan                          & Mikoyan mig-29                             & Milkfish                           \\
Milling                   & Mitsuoka viewt                  & Motor torpedo boat                & Motorcycle speedway                        & Musk deer                          \\
Muskox                    & Nebelung                        & Nectar                            & Negroni                                    & Network interface controller       \\
Newt                      & Nightingale                     & Norwegian forest cat              & Norwegian lundehund                        & Nova scotia duck tolling retriever \\
Obstacle race             & Offshore drilling               & Oil field                         & Omelette                                   & Orangery                           \\
Ortolan bunting           & Patriarch                       & Patrol boat                       & Patterdale terrier                         & Pda                                \\
Peccary                   & Peppers                         & Perch                             & Persian                                    & Pilot boat                         \\
Pinscher                  & Piper pa-18                     & Pit cave                          & Pixie-bob                                  & Platform supply vessel             \\
Player piano              & Polish hunting dog              & Polish lowland sheepdog           & Pomacentridae                              & Pontiac chieftain                  \\
Porsche 911 gt2           & Porsche 911 gt3                 & Porsche 959                       & Portuguese water dog                       & Potato pancake                     \\
Pronghorn                 & Pumi                            & Quesadilla                        & Quinceañera                                & Ragdoll                            \\
Rattlesnake               & Reefer ship                     & Republic p-47 thunderbolt         & Requiem shark                              & Research vessel                    \\
Rice ball                 & Rings                           & Roller hockey                     & Roller in-line hockey                      & Rolls-royce phantom                \\
Rolls-royce phantom iii   & Rowan                           & Lugger                            & Ruf ctr                                    & Russell terrier                    \\
Russian blue              & Rusty-spotted cat               & Saarloos wolfdog                  & Sailboat racing                            & Salep                              \\
Salisbury steak           & Salt-cured meat                 & Samgyeopsal                       & Sardine                                    & Savannah                           \\
Schisandra                & Schnecken                       & Shenyang j-11                     & Shiba inu                                  & Shikoku                            \\
Ship of the line          & Sighthound                      & Sikorsky s-61                     & Ski cross                                  & Ski touring                        \\
Skyway                    & Slalom skiing                   & Slide guitar                      & Sloppy joe                                 & Smooth newt                        \\
Snipe                     & Snowshoe                        & Sport climbing                    & Sport kite                                 & Stabyhoun                          \\
Stall                     & Steak tartare                   & Submarine chaser                  & Sukhoi su-27                               & Sukhoi su-30mkk                    \\
Sukhoi su-35bm            & Synagogue                       & Taekkyeon                         & Tamaskan dog                               & Tapestry                           \\
Telemark skiing           & Tervuren                        & Tibetan spaniel                   & Tiki                                       & Torpedo boat                       \\
Toy fox terrier           & Toyger                          & Trout                             & Tteokbokki                                 & Tufted capuchin                    \\
Tvr                       & Ukulele                         & Uneven bars                       & Vienna sausage                             & Vihuela                            \\
Viol                      & Virginia opossum                & Viverridae                        & Volkswagen golf mk5                        & Volkswagen type 14a                \\
Volpino italiano          & Vought f4u corsair              & Wasserfall                        & Roller in-line hockey                      & Weed                               \\
Western pleasure          & White shepherd                  & World rally championship          & Wren                                       & Yarrow                             \\
Zwiebelkuchen             & Épée                            &                                   &                                            &                                    \\
\bottomrule
\end{tabular}
\end{table*}

\end{document}